\documentclass{article}

\PassOptionsToPackage{numbers, compress}{natbib}

\usepackage[preprint]{neurips_data_2022}

\usepackage[utf8]{inputenc} 
\usepackage[T1]{fontenc}    
\usepackage{hyperref}       
\usepackage{url}            
\usepackage{booktabs}       
\usepackage{amsfonts}       
\usepackage{nicefrac}       
\usepackage{microtype}      
\usepackage{xcolor}         
\usepackage[pdftex]{graphicx}
\usepackage{tabularx}
\usepackage{pifont}
\usepackage{multirow}
\usepackage{wrapfig}
\usepackage{rotating}
\usepackage{floatrow}
\usepackage{blindtext}
\usepackage{titlesec}
\usepackage{float}

\newfloatcommand{capbtabbox}{table}[][\FBwidth]

\newcommand{\cmark}{\ding{51}}%
\newcommand{\dataset}{Multimodal Lecture Presentations Dataset}
\newcommand{\datasets}{Multimodal Lecture Presentations Dataset }
\newcommand{\datasetshort}{MLP Dataset}
\newcommand{\datasetshorts}{MLP Dataset }

\newcommand{\model}{PolyViLT}

\newcommand{\weburl}{\url{https://github.com/dondongwon/MLPDataset}}

\title{\dataset: Understanding Multimodality in Educational Slides}

\author{%
    Dong Won Lee, Chaitanya Ahuja, Paul Pu Liang, Sanika Natu, Louis-Philippe Morency   \\
    Carnegie Mellon University\\
    \weburl
}

\begin{document}

\maketitle

\vspace{-6mm}
\begin{abstract}

Lecture slide presentations, a sequence of pages that contain text and figures accompanied by speech, are constructed and presented carefully in order to optimally transfer knowledge to students. Previous studies in multimedia and psychology attribute the effectiveness of lecture presentations to their multimodal nature. As a step toward developing AI to aid in student learning as intelligent teacher assistants, we introduce the Multimodal Lecture Presentations dataset as a large-scale benchmark testing the capabilities of machine learning models in multimodal understanding of educational content. Our dataset contains aligned slides and spoken language, for 180+ hours of video and 9000+ slides, with 10 lecturers from various subjects (e.g., computer science, dentistry, biology). We introduce two research tasks which are designed as stepping stones towards AI agents that can \textit{explain} (automatically captioning a lecture presentation) and \textit{illustrate} (synthesizing visual figures to accompany spoken explanations) educational content. We provide manual annotations to help implement these two research tasks and evaluate state-of-the-art models on them. Comparing baselines and human student performances, we find that current models struggle in (1) weak crossmodal alignment between slides and spoken text, (2) learning novel visual mediums, (3) technical language, and (4) long-range sequences. Towards addressing this issue, we also introduce \model, a multimodal transformer trained with a multi-instance learning loss that is more effective than current approaches. We conclude by shedding light on the challenges and opportunities in multimodal understanding of educational presentations.

\end{abstract}


\vspace{-5mm}
\section{Introduction}
\vspace{-1mm}


Students today commonly learn through multimedia, including online lecture presentation recordings, educational mobile applications, and other digital resources~\cite{mayer2002multimedia}. In particular, slide-assisted instruction through lectures has become predominant in educational settings~\cite{piolat2005cognitive,reedy2008powerpoint,savoy2009information} and is widely considered by teachers and students as the preferred instructional tool~\cite{reedy2008powerpoint,susskind2005powerpoint}. The effectiveness of lecture slides is supported by research in multimedia principles, which show that individuals learn more effectively from spoken (or written) language when  accompanied by graphics rather than language in isolation~\cite{baddeley2003working,mayer1991animations,mayer2002aids,moreno1999cognitive,paivio1990mental}. The prevalence and effectiveness of lecture slides as an educational medium calls for AI systems that are also able to understand and communicate multimodal knowledge, in order to move closer towards intelligent teaching assistants \cite{griol2013architecture}. More specifically, AI systems that can understand lecture slides could yield many exciting applications, such as an intelligent tutoring system that retrieves a slide to answer a student's question, a recommender system that automatically generates a slide on-the-fly as the speaker is speaking, or an evaluation system that evaluates the quality of lecture presentations.

As a step towards this direction, we design the \datasets\ (\datasetshort) as a large-scale benchmark evaluating AI technologies in multimodal understanding of educational content. \datasetshort\ contains over 9000 slides with natural images, diagrams, equations, tables and written text, aligned with the speaker's spoken language. These lecture slides are sourced from over 180 hours worth of educational videos in various disciplines such as anatomy, biology, psychology, speaking, dentistry, and machine learning. To benchmark the understanding of multimodal information in lecture slides, we introduce two research tasks which are designed to be a first step towards developing AI that can explain and illustrate lecture slides: automatic retrieval of (1) spoken explanations for an educational figure (Figure-to-Text) and (2) illustrations to accompany a spoken explanation (Text-to-Figure)\footnote{Text-to-Figure can be thought as a recommender system assisting a lecturer by providing recommendations of figures when building presentation slides according to the lecturer's planned transcript. On the other hand, Figure-to-Text can be used by an intelligent tutoring system to retrieve relevant explanations to help a student when given a query of a visual diagram.}. To enable these tasks, we manually annotated the slide segments to accurately capture alignment between spoken language, slides, and figures (diagrams, natural images, table, equations).

\datasetshorts and its tasks bring new research opportunities through the following technical challenges: (1) addressing weak crossmodal alignment between figures and spoken language (a figure on the slide is often related to only a portion of spoken language), (2) representing novel visual mediums of man-made figures (e.g., diagrams, tables, and equations), (3) understanding technical language, and (4) capturing interactions in long-range sequences. Through human and quantitative studies, we find that current multimodal models struggle with the aforementioned challenges. We work towards addressing weak alignment and novel visual mediums by introducing \model, a multimodal transformer trained with a multi-instance learning loss. Although \model\ presents some improvement, \datasetshorts still offers novel challenges that will spark future research in educational content modeling, multimodal reasoning, and question answering.

\vspace{-2mm}
\section{Related Work}
\vspace{-2mm}

\begin{figure*}[t]
    \begin{center}
    \vspace{-4mm}
    \makebox[\textwidth][c]{\includegraphics[width=1.2\textwidth]{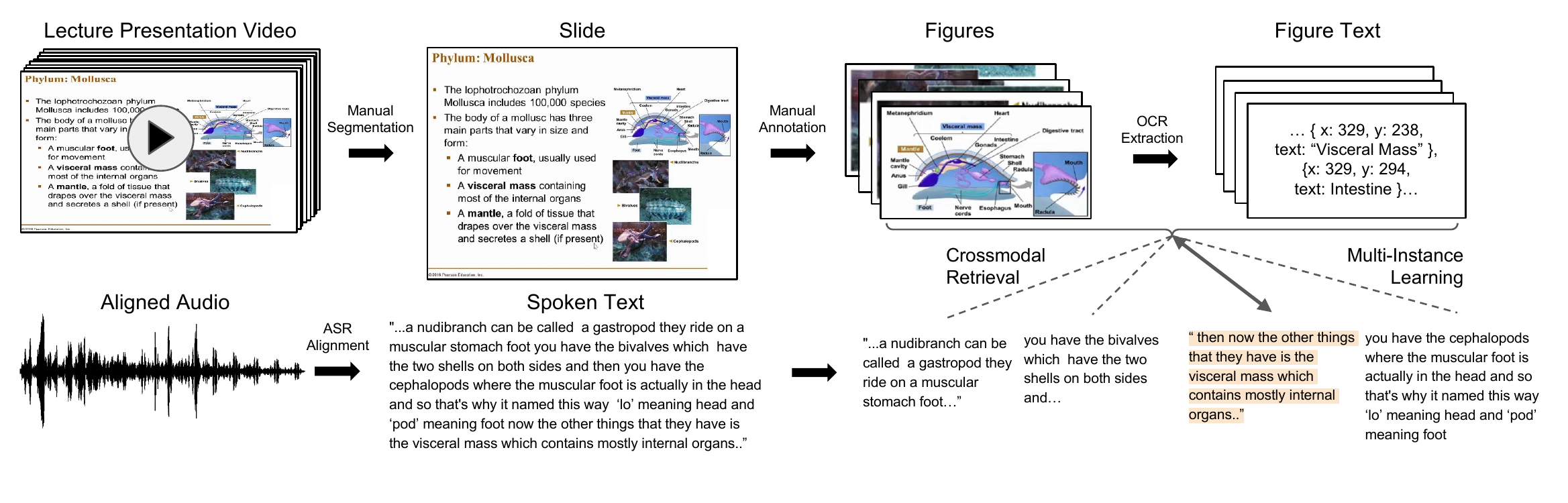}}
    \vspace{-10mm}
    \caption{We present the \datasets as benchmark for developing AI technologies that can communicate multimodal knowledge in educational content. Our diversely sourced and richly annotated dataset contributes two challenging research tasks: automatic retrieval of (1) spoken explanations given figures and (2) illustrative figures given spoken explanations. Through benchmarking existing and newly proposed models, we outline future research directions in tackling weak crossmodal alignment, novel visual mediums, technical language, and long-range sequences to bring us closer to intelligent and accessible tutoring aids.\vspace{-2mm}}
    \label{fig:main}        
    \end{center}
\end{figure*}


The effectiveness of lecture slides as a medium of transferring information can be explained by five multimedia learning principles \cite{garner2013design}. Firstly, the multiple representation principle states that individuals learn more effectively from graphics accompanied by spoken or written verbal information than solely spoken language. This principle is supported by dual-route processing mechanisms of working memory aand comprehension processes, where integration of verbal and nonverbal information benefits formation of representations in working memory. \cite{mayer2002aids, mayer1991animations, paivio1990mental, baddeley2003working, moreno1999cognitive}. Secondly, the contiguity principle expounds on reducing the spatio-temporal separation between different forms of information, which decreases the amount of effort required to build a coherent mental representation \cite{chandler1991cognitive, paas2004cognitive}. Third, redundancy, the exposure to complementary but identical information in different modalities, improves learner's working memory (auditory, visual). Fourth, coherence: restricting the information presented to only essential information allows the learner to integrate key concepts and relationships \cite{harp1997role}. Finally, signalling provides learner information regarding the overall hierarchical structure of the presentation \cite{mayer2002aids}.

\begin{table}[ht]
\resizebox{1\textwidth}{!}{%
\begin{tabular}{@{}l|llll|lll|l@{}}
\toprule
 & \multicolumn{4}{c|}{\textbf{Features}} & \multicolumn{3}{c|}{\textbf{Size}} & \multicolumn{1}{c}{\textbf{Avail.}} \\ \midrule
 & \multicolumn{1}{l|}{Slide Segments} & \multicolumn{1}{l|}{Slide Figures} & \multicolumn{1}{l|}{Slide Text} & Spoken Language & \multicolumn{1}{c|}{\# Videos} & \multicolumn{1}{c|}{\# Hours} & \multicolumn{1}{c|}{\# Slides} & \multicolumn{1}{c}{} \\ \midrule
VLEngagment \cite{bulathwela2020vlengagement} & \multicolumn{1}{l|}{} & \multicolumn{1}{l|}{} & \multicolumn{1}{l|}{} &  & \multicolumn{1}{l|}{11568} & \multicolumn{1}{l|}{} &  & \cmark \\
LectureBank \cite{li2019should} & \multicolumn{1}{l|}{\cmark (M)} & \multicolumn{1}{l|}{} & \multicolumn{1}{l|}{\cmark (A)} &  & \multicolumn{1}{l|}{1352} & \multicolumn{1}{l|}{} & 51,939 & \cmark \\
ALV \cite{galanopoulos2019temporal} & \multicolumn{1}{l|}{\cmark (A)} & \multicolumn{1}{l|}{} & \multicolumn{1}{l|}{} & \cmark (A) & \multicolumn{1}{l|}{} & \multicolumn{1}{l|}{} & 1498 & \cmark \\
LectureVideoDB \cite{lectureVideoDB2018} & \multicolumn{1}{l|}{\cmark (M)} & \multicolumn{1}{l|}{} & \multicolumn{1}{l|}{\cmark (M)} &  & \multicolumn{1}{l|}{24} & \multicolumn{1}{l|}{} & 5000 & \cmark \\
GoogleI/O \cite{chen2014multi} & \multicolumn{1}{l|}{} & \multicolumn{1}{l|}{} & \multicolumn{1}{l|}{\cmark (A)} & \cmark (A) & \multicolumn{1}{l|}{209} & \multicolumn{1}{l|}{174} &  & \cmark \\
LaRochelle \cite{van2014multi} & \multicolumn{1}{l|}{\cmark (A)} & \multicolumn{1}{l|}{} & \multicolumn{1}{l|}{\cmark (A)} & \cmark (A) & \multicolumn{1}{l|}{47} & \multicolumn{1}{l|}{65} & 2350 &  \\
\datasetshort & \multicolumn{1}{l|}{\cmark (M)} & \multicolumn{1}{l|}{\cmark (M)} & \multicolumn{1}{l|}{\cmark (A)} & \cmark (A) & \multicolumn{1}{l|}{334} & \multicolumn{1}{l|}{187} & 9031 & \cmark \\ \bottomrule
\end{tabular}%
}
\vspace{-1mm}
\caption{Comparison with existing lecture-based datasets, (A) represents automatic processing, (M) represents manual processing. MLP Dataset is the first of its kind to offer slide segmentation, aligned spoken language, slide text, and visual figures, while being publicly available.\vspace{-4mm}}
\label{fig: related_data}
\end{table}

\vspace{-1mm}
Given the effectiveness of lecture slides as a medium of presenting information, future AI should be able to learn and extract from the carefully curated, rich information in lecture slides. Recent work towards computational modeling of lecture slides include LectureBank \cite{fabbri2018tutorialbank,li2019should}, ALV \cite{galanopoulos2019temporal}, VLEngagement \cite{bulathwela2020vlengagement}, LectureVideoDB \cite{lectureVideoDB2018}, GoogleI/O \cite{chen2014multi}, and LaRochelle \cite{van2014multi}. We summarize and compare these datasets with our proposed \datasetshort\ in Table \ref{fig: related_data}. We also highly recommend readers to refer to Appendix \ref{asec:prev_lec} where we present detailed descriptions and differences of previous datasets. To the best of our knowledge, \datasetshort\ is the first of its kind to offer slide segmentation, aligned spoken language, slide text, and visual figures, while being publicly available for the research community.

\vspace{-2mm}
\section{\datasets}
\vspace{-2mm}
The \datasets\ is designed as a benchmark to develop AI models capable of understanding multimodal information present in lecture slides. Our dataset offers segmented slides, their aligned spoken language and visual elements (figures, diagrams, natural images, tables), and OCR text output for slide text.

\vspace{-2mm}
\subsection{Problem Definition}
\vspace{-2mm}

In order to benchmark an AI technology's understanding of multimodal education content and move closer to automatic generation of captions and figures, we measure its ability to associate a visual figure with a spoken explanation. We define figures as a visual illustration consisting of text, images, drawings (i.e., diagrams, images, equations, or tables). We design 2 proxy tasks, where (1) visual figures are retrieved given spoken language (Text-to-Figure) and (2) spoken explanations are retrieved from visual figures (Text-to-Figure). In contrast to many prior crossmodal retrieval setups which assume one-to-one mappings between modalities \cite{wang2016comprehensive}, lecture presentations are unique in the presence of weak crossmodal alignment between spoken language and figures. There could exist $n > 1 $ visual figures for a single spoken speech segment $s$ and a figure could be aligned partially to the spoken segment (i.e., a part of the spoken segment is used to explain the figure). Thus, a core challenge lies in addressing weak crossmodal alignment. Formally, let $D = (S,V)$ be our dataset consisting of spoken language $S$ and figures $V$. The goal is to learn an embedding space that can quantify the similarity between the figure and spoken language. As a result, given a segment of spoken language $s \in S$,  one could retrieve the set of aligned visual figures $\{ v_k, v_{k+1}, ...,v_{k+n} \} \subseteq V$.

 
\vspace{-2mm}
\subsection{Dataset Statistics}
\vspace{-2mm}

\begin{figure*}[t]
    \begin{center}
    \makebox[0.8\textwidth][c]{\includegraphics[width=1\textwidth]{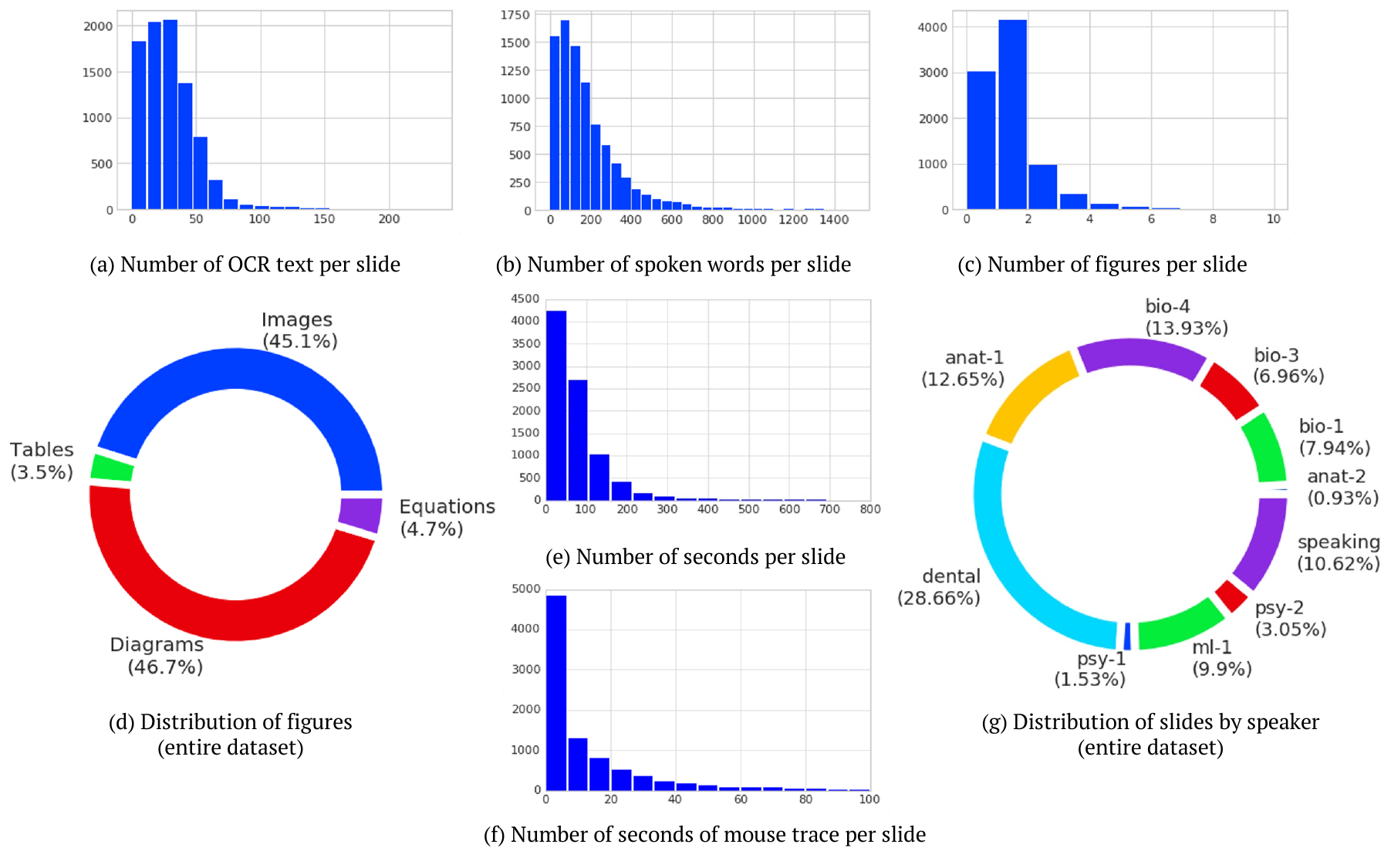}}
    \caption{Dataset statistics for \dataset. 2(d) shows that our a large majority of our dataset consists of figures that include text (diagrams, equations, tables). Figure 2(g) shows that the variety of educational content covered by our dataset.\vspace{-4mm}}
    \label{fig:dataset_stats}
    \end{center}
\end{figure*}

Our \datasetshort\ consists of 9031 slides, 8598 figures, 28000 unique words, 1.6 million total words from 334 educational presentation videos with a total duration of 187 hours. As shown in Table \ref{fig:dataset_stats} (b) Per slide, there are 185.83 words spoken on average and 135 words in median. Each slide's duration is an average of 72.6 seconds or a median of 54.9 seconds as shown in Figure Table \ref{fig:dataset_stats}(e). Among the 8598 figures,  3877 (45.1\%) are natural images, 4018 (46.7\%) are diagrams, 301 (3.5\%) are tables, 402 (4.6\%) are equations, shown in Table \ref{fig:dataset_stats}(d). Each slide has a median of 1 (or 0.94 on average) figure, shown in Table \ref{fig:dataset_stats}(c), a median of 26 written words (or 28.95 on average), displayed in Table \ref{fig:dataset_stats}(a). Furthermore, when available, we provide the mouse traces which hover over the region the speaker is describing. There are 12.52 seconds of mouse trace data per slide on average and 4.6 in median, as shown in Table \ref{fig:dataset_stats}(f). Our dataset consists of 35 courses on biology, anatomy, psychology, dentistry, speaking, machine learning taught by 10 speakers. The distribution of the number of slides per speaker is shown in Table \ref{fig:dataset_stats}(g).



\vspace{-2mm}
\subsection{Data Collection and Preprocessing}
\label{sec:data_proc}
\vspace{-2mm}

\begin{figure*}[t]
    \begin{center}
    \makebox[\textwidth][c]{\includegraphics[width=1.2\textwidth]{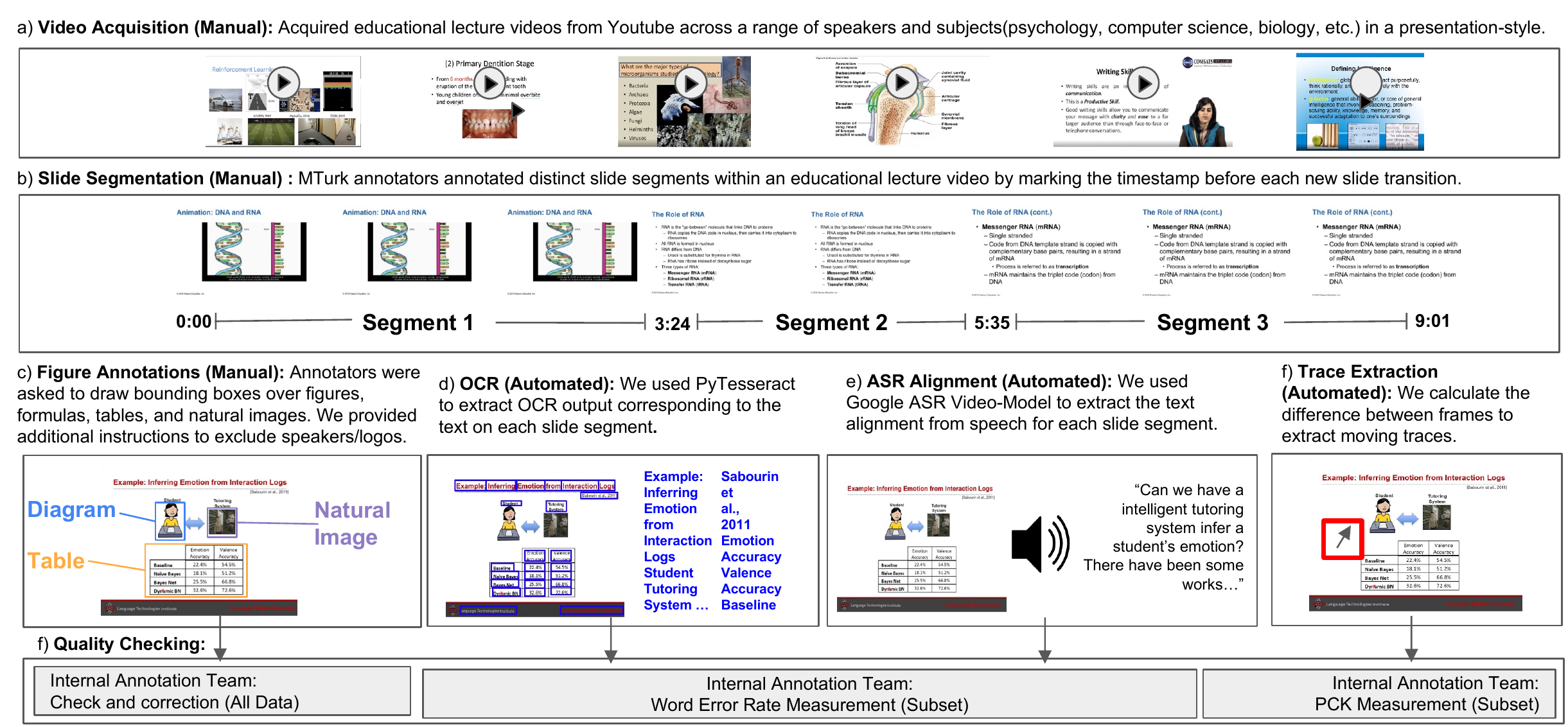}}
    \caption{Overview of our data collection and preprocessing steps with a summary of each step. Best viewed zoomed in and in color.\vspace{-4mm}}
    \label{fig:dataset_col}
    \end{center}
\end{figure*}

The \datasets is developed from a curated list of lecture presentation videos, which are downloaded from YouTube. Spoken language is then extracted from speech via automatic speech recognition. We manually annotate for the slide segments as well as figure bounding boxes and corresponding labels in order to perform retrieval tasks between slide-level segments spoken text and individual visual figures. In addition, in order to utilize the language information in figures, the texts in the slide are automatically extracted via OCR. A visual outline the data collection and processing steps taken to create \datasetshorts is shown in Figure \ref{fig:dataset_col}. 

\vspace{-1mm}
\textbf{Video Acquisition}: 413 English educational videos were downloaded from YouTube. From the initial list, we filtered and curated a smaller list of 10 speakers according to the following criteria: (1) the material must be presented in a slide-based style, (2) the slides must be stationary (i.e. external video clips cannot be played), and (3) the speaker makes use of their mouse to refer to specific figures on the slide. After filtering, 334 videos remained.

\vspace{-1mm}
\textbf{Slide Segmentation}:
The quality of segmentation is crucial to our task of retrieval, therefore, we collected manual human annotations on MTurk. We presented the annotator with a lecture video and asked the annotator to use a slider to navigate to the end of each slide and mark its precise timestamp. A screenshot of this experiment can be found in Appendix \ref{asec:mturk_slideseg}. 

In order for ensure the high quality of segmentations, we conduct the annotation process in multiple steps. (1) An internal team manually annotated 10 lecture videos for groundtruth annotations. (2) The experiment was made available to 100 MTurkers. (3) We evaluate their results, marking an annotation as correct if theirs matched ours within a 1 second interval. (4) Annotators who were able to perform above a 90\% correctness threshold were assigned the full set.

\vspace{-1mm}
\textbf{Figure Annotation and Labeling}:
Our dataset is unique from previous datasets as our focus is centered around figure-level retrieval. In order to enable this task, our data must consist of precise bounding boxes and labels for each figure. Therefore, we design an MTurk experiment where annotators are shown slides and asked to create a bounding box around figure instances and label their classes. Our class labels are inspired from PRImA \cite{antonacopoulos2009realistic}, a dataset that consists of layouts from scientific reports. We follow their taxonomy to find labels on figures, which consist of natural images, diagrams, table, and equations. In Appendix \ref{asec:mturk_figanot}, we provide details on figure class labels and a screenshot of the MTurk experiment.

To obtain precise and accurate figure annotations, we follow a multi-phase process. (1) An internal team manually annotated 10 lecture videos for groundtruth figure annotations and labels. (2) We make the experiment available to 100 MTurkers for 10 different slides. (3) We manually evaluate the annotations, marking an annotation as correct if the annotators had the same number of figures, equivalent types, and high overlap of bounding boxes. (4) Annotators who were able to perform above a 90\% correctness threshold were assigned the full set. (5) To ensure the absolute highest quality of figure annotations, our internal team of annotators manually corrected all the annotations for any mislabeled bounding box annotations or incorrect regions.




\vspace{-1mm}
\textbf{Text Extraction: ASR \& OCR}:
We use Google ASR \cite{chiu2018state} to extract spoken language from audio. We use the Video-Model, which has a reported WER of 16\% (Amazon: 22\%, Microsoft 24\%, IBM Watson 29\%, Google Speech-Model 37\%). We manually verify 100 random segments in the dataset, and find that the WER is 17.1\%. To extract OCR text from the images of slides, we use Tesseract \cite{smith2007overview}. We manually verify 100 random slides in the dataset, and find that the WER is 37.82\%.

\vspace{-1mm}
\textbf{Mouse Trace Location Extraction}:
We extract the mouse trace location to be used as an additional grounding signal between visual objects and language. For each segmented slide, the background is static and the only object that is moving is the pointer. If there is any movement, we consider that as the pointer location. We manually verify 100 random mouse trace location in the dataset, and find that the percentage of correct keypoints (PCK) with a threshold of 50 pixels, is 77.1\%.


\vspace{-3mm}

\begin{figure*}[t]
    \begin{center}
    \vspace{-2mm}
    \caption{ Comparison of baselines and PolyViLT against human student performance in Recall@10 for (Top) Figure-to-Text and (Bottom) Text-to-Figure retrieval. }
    \makebox[\textwidth][c]{\includegraphics[width=1.2\textwidth]{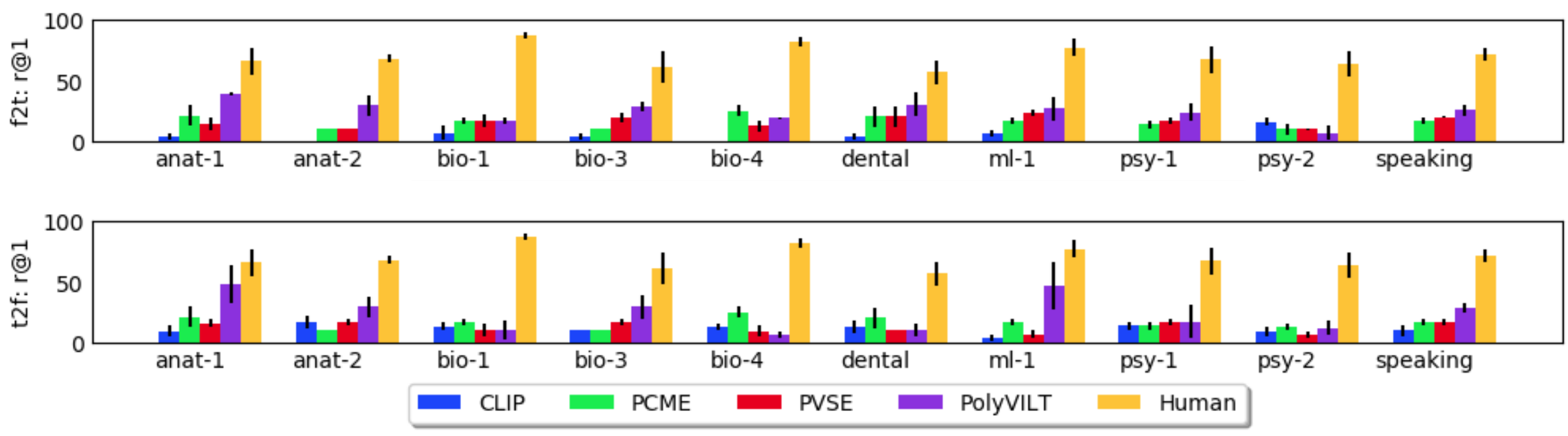}}
    \label{fig:i2t_human}
    \vspace{-4mm}
    \end{center}
\end{figure*}

\section{Experimental Setup}
\vspace{-3mm}

The \datasetshort is designed to examine multimodal model's understanding of educational material, as measured by its performance on text-to-figure and figure-to-text retrieval. We evaluate multiple state-of-art model's performance in comparison with human student performance. We are interested in understanding how current state-of-the-art models perform on different figure types (diagrams, images, equations, tables), long range sequences, and technical language. We also introduce PolyViLT, a multi-instance learning multimodal transformer that utilizes both vision and language information in slide figures.


\vspace{-2mm}
\subsection{Baselines}
\label{sec:baselines}
\vspace{-2mm}


We select previous baselines PVSE \cite{song2019polysemous} and PCME \cite{chun2021probabilistic} that are designed for cross-modal retrieval particularly in scenarios with weak alignment. We also measure CLIP \cite{radford2021learning} performance, as its zero-shot image-text matching performance is well recognized in the community. 

\vspace{-1mm}
\noindent \textbf{CLIP \cite{radford2021learning}} is an established baseline for image-text matching.
We use a pre-trained CLIP model to embed pairs of figures and text and rank according to their similarity scores for retrieval.

\vspace{-1mm}
\noindent \textbf{PVSE \cite{song2019polysemous}} is designed to model one-to-many alignment for crossmodal retrieval, by encoding visual and text features as $K$ possible embeddings and training with a multiple instance loss that rewards weak cross-modal alignment (i.e., the best pair among $K^2$ pairs is rewarded).

\vspace{-1mm}
\noindent \textbf{PCME \cite{chun2021probabilistic}} handles pairwise semantic similarities and uncertainty in crossmodal retrieval. It models each modality as probabilistic distributions in a common embedding space using Hedged Instance Embeddings (HIB) \cite{oh2018modeling} and utilizes a soft version of the contrastive loss to handle weak alignment.

\vspace{-2mm}
\subsection{PolyViLT: A Proposed Model for Weak Image-Text Alignment}
\label{sec:polyvilt}
\vspace{-2mm}

On top of these baselines, we further introduce Polysemous-ViLT (or PolyViLT), which is designed to handle vision and language inputs (e.g., diagrams) and weak cross-modal alignment. Previous approaches were designed specifically for the task of crossmodal retrieval on datasets consisting of only natural images and text. However, to perform well on retrieval problems involving figures, models must utilize text information present in the figure, as they could provide valuable signals to the model. Our approach utilizes local feature transformers in PVSE \cite{song2019polysemous}, a multi-instance learning loss \cite{dietterich1997solving} and a ViLT figure encoder \cite{kim2021ViLT} to utilize both vision and language information in figures. We refer the readers to Figure \ref{fig:model} for details and figure of the model architecture.

\noindent \textbf{ViLT Figure Encoder} We utilize the ViLT model \cite{kim2021ViLT} as a backbone encoder to contextualize the text and vision information present in figure. Given an image of a figure, the accompanying text (from OCR output) on the figure is tokenized with BERT \cite{devlin2018bert}, patches of the diagram image is flattened and linearly projected, and fed in as a sequence to a transformer encoder. We initialize the ViLT encoder with pretrained weights trained on masked language modelling and image-text matching before training on our dataset. 

\noindent \textbf{Multiple Instance Learning (MIL)} To account for the partial alignment between figures and spoken language, we represent the spoken language with $K$ embeddings, capturing different words of the speech, inspired by local feature transformers in \cite{song2019polysemous}, The local $K$ embedding are combined with global information via residual connections. Then, we utilize the MIL objective \cite{dietterich1997solving}, which assume that there is a partial match between a figure and $K$ local embeddings of the spoken language. 

\vspace{-2mm}
\subsection{Human Student Performance}
\vspace{-2mm}
\label{ssec:human_perf}

To measure human student performance, we randomly sampled 10 figures from the unseen test set for each speaker from 3 random seeds. For Figure-to-Text, a student is shown one figure image, all the spoken language aligned to the 10 figures in the sample and is asked to select the most relevant spoken language. For Text-to-Figure, the annotator is shown one spoken language, all the figure and is asked to select the most relevant figure. We report recall@1 metric for this sample for all of our baseline models for fair comparison.



\vspace{-2mm}
\section{Results}
\vspace{-2mm}

\begin{figure*}[t]
    \vspace{-0mm}
    \begin{center}
    \makebox[\textwidth][c]{\includegraphics[width=1\textwidth]{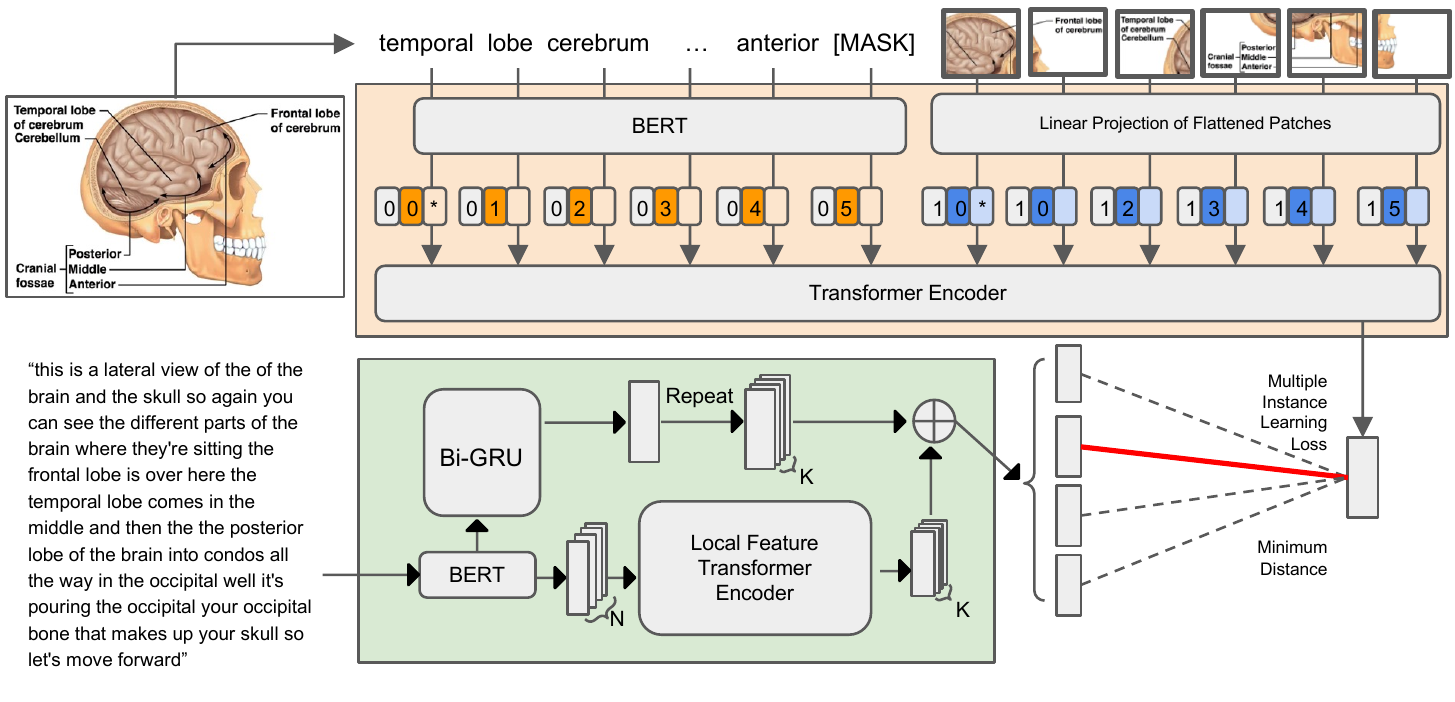}}
    \caption{Overview of the key components of our proposed PolyViLT model. The diagram's text and image patches are input into a ViLT-based transformer encoder, and the spoken language BERT embedding is transformed into $K$ representations. A MIL Loss \cite{dietterich1997solving} is used to address weak crossmodal alignment and find partially aligned instances.\vspace{-2mm}}
    \label{fig:model}
    \end{center}
\end{figure*}

\vspace{-2mm}
\subsection{Model and Human Performance}
\vspace{-2mm}

The performance of all models can be seen in Table \ref{tab: main}. PolyViLT outperforms previous state-of-the art approaches in both Figure-to-Text retrieval and Text-to-Figure Retrieval. The second best performing model is PVSE \cite{song2019polysemous}, which further justifies our reasoning behind utilizing local feature transformers and the MIL loss. Surprisingly, CLIP's zero-shot performance often is worse than Random, which indicates that large-scale pre-training on natural image-text pairs may not be sufficient for our task. The detailed results for each speaker can be found in Appendix \ref{asec:each_sp_results}. We also provide human student retrieval performance in Figure~\ref{fig:i2t_human}. We see that all methods fall well below human students' performance, even PolyViLT, the closest method, is 47.68\% worse for Text-to-Figure retrieval and 43.63\% worse for Figure-to-Text retrieval, which demonstrates the challenging nature of our dataset. In the following sections, we perform error analysis to uncover the concrete challenges presented in \datasetshort.

\vspace{-4mm}
\subsection{Performance on Novel Visual Mediums}
\vspace{-2mm}

We first investigate the impact of novel visual mediums such as man-made figures (e.g., diagrams, tables, and equations) on model performance.
We report recall@10 scores conditioned on each type in Table \ref{tab:res_by_fig}, and find that PolyViLT outperforms other baselines for most figure types. Interestingly, we can see that for natural images, previous approaches perform worse than PolyViLT. Whereas we believed that PolyViLT's main advantage is in its use of text information, it outperforms previous approaches even when no text information is used. This indicates that the usage of a ViT encoder \cite{dosovitskiy2020image} is superior over using local and global feature transformers as proposed in PVSE \cite{song2019polysemous} and PCME \cite{chun2021probabilistic} even for natural images. We also find models struggle particularly on equations. As mentioned in Section \ref{sec:polyvilt} this could be attributed to the significant domain difference between the pretraining domain (natural images, non-educational language) of ViLT \cite{kim2021ViLT} and equations. PVSE \cite{song2019polysemous} is initialized with random weights, therefore is unaffected.  

\begin{table}
\resizebox{\textwidth}{!}{%
\begin{tabular}{@{}llllllll@{}}
\toprule
\multirow{2}{*}{\textbf{Models}} & \multicolumn{3}{c}{\textbf{Figure-to-Text}} &  & \multicolumn{3}{c}{\textbf{Text-to-Figure}} \\ \cmidrule(lr){2-4} \cmidrule(l){6-8} 
 & Recall@1 & Recall@5 & Recall@10 &  & Recall@1 & Recall@5 & Recall@10 \\ \midrule
Random & 1.36 ± 0.22 & 7.63 ± 0.88 & 15.81 ± 0.7 &  & 2.15 ± 0.61 & 8.64 ± 1.1 & 16.38 ± 1.91 \\
CLIP \cite{radford2021learning} & 2.05 ± 0.7 & 7.4 ± 0.15 & 17.65 ± 1.02 &  & 1.58 ± 0.56 & 6.89 ± 1.18 & 13.78 ± 0.55 \\
PVSE \cite{song2019polysemous} & 3.17 ± 0.68 & 12.44 ± 1.28 & 22.01 ± 0.61 &  & 2.81 ± 0.27 & 11.87 ± 1.24 & 21.2 ± 0.63 \\
PVSE (BERT) \cite{song2019polysemous, devlin2018bert} & 2.96 ± 0.76 & 10.96 ± 0.52 & 18.54 ± 0.99 &  & 2.43 ± 0.05 & 11.21 ± 1.11 & 18.51 ± 1.1 \\
PCME \cite{chun2021probabilistic} & 2.31 ± 0.41 & 8.83 ± 0.34 & 16.43 ± 0.67 &  & 2.12 ± 0.36 & 8.68 ± 0.14 & 16.9 ± 1.1 \\
PCME (BERT) \cite{chun2021probabilistic, devlin2018bert} & 1.93 ± 0.26 & 8.27 ± 0.95 & 15.76 ± 1.64 &  & 1.93 ± 0.26 & 8.36 ± 1.08 & 15.85 ± 1.77 \\ \midrule
PolyViLT + Trace & 3.85 ± 0.91 & 17.77 ± 1.88 & 28.26 ± 1.78 &  & 5.38 ± 0.78 & 19.66 ± 2.39 & 32.26 ± 0.59 \\
\textbf{PolyViLT} & \textbf{4.94 ± 0.55} & \textbf{19.16 ± 0.69} & \textbf{30.35 ± 0.55} &  & \textbf{6.14 ± 1.25} & \textbf{23.19 ± 0.68} & \textbf{33.22 ± 1.73} \\ \bottomrule
\end{tabular}%
}
\caption{Comparison between PolyViLT vs previous state-of-the-art models for crossmodal retrieval with multiple instance learning across all dataset for 3 random seeds, standard deviation bars are reported. PolyViLT outperforms all previous state-of-the-art approaches by a large margin.\vspace{-2mm}}
\label{tab: main}
\end{table}
\begin{table}
\vspace{-0mm}
\resizebox{\textwidth}{!}{%
\begin{tabular}{@{}llllllllll@{}}
\toprule
\multirow{2}{*}{\textbf{Models}} & \multicolumn{4}{c}{\textbf{Figure-to-Text:  Recall@10}} &  & \multicolumn{4}{c}{\textbf{Text-to-Figure: Recall@10}} \\ \cmidrule(lr){2-5} \cmidrule(l){7-10} 
 & Diagram & Image & Table & Equation &  & Diagram & Image & Table & Equation \\ \midrule
CLIP \cite{radford2021learning} & 6.2 ± 0.57 & 5.77 ± 0.73 & 6.2 ± 4.36 & 2.83 ± 1.11 &  & 6.5 ± 1.27 & 6.0 ± 0.22 & 6.9 ± 2.5 & 3.5 ± 0.96 \\
PVSE \cite{song2019polysemous} & 8.2 ± 0.93 & 9.6 ± 0.57 & 7.27 ± 0.29 & \textbf{12.27 ± 3.27} &  & 7.6 ± 1.3 & 10.33 ± 1.76 & 6.97 ± 4.15 & 4.47 ± 4.66 \\
PCME \cite{chun2021probabilistic} & 6.0 ± 0.37 & 6.9 ± 0.22 & 6.3 ± 3.28 & 2.93 ± 3.27 &  & 5.9 ± 0.49 & 6.87 ± 0.26 & 6.3 ± 3.28 & 2.93 ± 3.27 \\ \midrule
\textbf{PolyViLT} & \textbf{18.53 ± 1.65} & \textbf{15.2 ± 0.91} & \textbf{15.83 ± 2.67} & 5.53 ± 5.37 &  & \textbf{18.53 ± 1.89} & \textbf{20.13 ± 0.7} & \textbf{19.17 ± 6.34} & \textbf{9.97 ± 3.48} \\ \bottomrule
\end{tabular}%
}
\caption{Comparison of recall@10 scores for baselines conditioned on types of figures, mean and standard deviations are reported for 3 seeds across all speakers. PolyViLT outperforms previous baselines in most cases, except for equation text-to-figure retrieval.\vspace{-2mm}}
\label{tab:res_by_fig}
\end{table}


\vspace{-4mm}
\subsection{Technical Language and Long Range Sequences}
\vspace{-2mm}

The second challenge we investigate is the presence of technical language beyond commonly spoken and written text. Table \ref{tab:res_conditions}(b) shows the number of subwords tokenized by HuggingFace's BERT Tokenizer \cite{devlin2018bert, wolf2019huggingface}, which represents the number of Out-of-Vocabulary (OOV) tokens, a proxy measure for how much external knowledge is required to understand technical language. With an increasing number of subwords, there is a drop in performance, indicating that our models struggle to quickly acquire technical information or require external knowledge to perform well.

Furthermore, our dataset poses challenges in capturing information in long range language sequences  due to its educational nature. In Table \ref{tab:res_conditions}(a), we report recall@10 scores conditioned on the number of spoken words. PolyViLT's performance peaks between 100 and 200 words, and decreases with increasingly longer spoken phrases, or very short spoken phrases (under 100). This calls for a need to develop models for extremely long-range and short-range sequences. We refer the readers to Appendix \ref{asec:qualitative} where we display examples of instances where our current baselines fail when technical knowledge or understanding of long range interactions are required, and Appendix \ref{asec:language_problems} for the negative impacts of long range sequences and technical language on other baseline models.

\begin{table}[t]
\resizebox{\textwidth}{!}{%
\begin{tabular}{@{}llrrrlllrrr@{}}
\toprule
\begin{tabular}[c]{@{}l@{}}PolyViLT\\ r@10\end{tabular} & \multicolumn{5}{l}{\textbf{(a) Length of Spoken Language}} &  & \multicolumn{4}{l}{\textbf{(b) Number of Subwords}} \\ \cmidrule(l){2-11} 
 & \textless{}100 & \multicolumn{1}{l}{\textbf{100 - 200}} & \multicolumn{1}{l}{200 - 400} & \multicolumn{1}{l}{400 - 600} & 600+ &  & \textless{}\textbf{10} & \multicolumn{1}{l}{10 - 20} & \multicolumn{1}{l}{20-30} & \multicolumn{1}{l}{30 - 50} \\ \midrule
Figure-to-Text & \multicolumn{1}{r}{0.195} & \textbf{0.276} & 0.186 & 0.227 & 0.175 &  & \multicolumn{1}{r}{\textbf{0.218}} & 0.191 & 0.128 & 0.132 \\
Text-to-Figure & \multicolumn{1}{r}{0.177} & \textbf{0.280} & 0.207 & 0.186 & \multicolumn{1}{r}{0.14} &  & \multicolumn{1}{r}{\textbf{0.191}} & 0.156 & 0.136 & 0.124 \\ \bottomrule
\end{tabular}%
}
\vspace{0mm}
\caption{(a) PolyViLT performance drops for very short or very long sequences, (b) or with increasing number of subwords (technical terms).\vspace{-4mm}}
\label{tab:res_conditions}
\end{table}


\vspace{-2mm}
\subsection{Importance of MIL objective}
\vspace{-2mm}

\begin{table}
\resizebox{\textwidth}{!}{%
\begin{tabular}{@{}lllrcclc@{}}
\toprule
\multirow{2}{*}{} & \multicolumn{3}{c}{\textbf{Figure-to-Text}} &  & \multicolumn{3}{c}{\textbf{Text-to-Figure}} \\ \cmidrule(lr){2-4} \cmidrule(l){6-8} 
 & \multicolumn{1}{c}{bio-1} & \multicolumn{1}{c}{dental} & \multicolumn{1}{c}{ml-1} &  & bio-1 & \multicolumn{1}{c}{dental} & ml-1 \\ \midrule
No MIL & 26.18 ± 3.92 & 12.15 ± 1.67 & 12.58 ± 4.62 &  & 28.74 ± 1.32 & 12.32 ± 1.37 & 22.23 ± 3.83 \\
\textbf{MIL} & \textbf{31.29 ± 7.51} & \textbf{17.07 ± 1.66} & \textbf{19.32 ± 3.04} & \textbf{} & \textbf{32.24 ± 5.25} & \textbf{20.23 ± 0.12} & \textbf{24.84 ± 8.93} \\ \bottomrule
\end{tabular}%
\caption{Using MIL to handle weak crossmodal alignment leads to performance boosts.\vspace{-4mm}}
\label{tab:mil}
}
\end{table}

We investigate the effects of using a MIL objective to handle ambiguous alignment by comparing  PolyViLT with and without the MIL objective in Table \ref{tab:mil}. ``No MIL'' is the case where we optimize using the standard triplet ranking objective \cite{frome2013devise, kiros2014unifying}. Consistently, across all 3 speakers, we see that MIL is useful and leads to performance boosts by handling weak crossmodal alignment. 


\vspace{-2mm}
\subsection{Using Mouse Trace as a Grounding Signal}
\vspace{-2mm}

Finally, we experiment with utilizing mouse trace as an additional grounding signal to capture crossmodal alignment. With this intuition, we represent mouse traces as a one-hot vector with length equivalent to the spoken language sequence. For the indices corresponding to words when the mouse hovered over the figure, we assign it the value 1, indicating that the spoken word is directly aligned to the given figure and is conceptually similar to hard attention. We re-parameterize this categorical distribution with a Gumbel-Softmax \cite{jang2016categorical}, and use a dot-product attention with skip connections to fuse spoken language and mouse traces. The result for this model is shown in Table \ref{tab: main}, as `PolyViLT + Trace'. For certain speakers, the inclusion of mouse-trace data offers better performance. We refer the readers to Appendix \ref{asec:each_sp_results} for speaker-specific studies. Future work should aim at better utilizing the valuable information in mouse traces as a grounding signal~\cite{koh2021text,pont2020connecting}.

\vspace{-3mm}
\section{Discussion}
\vspace{-2mm}

\vspace{-1mm}
\textbf{Limitations}: Although our dataset presents exciting opportunities, it comes with its limitations. There exists an imbalance in slide distribution amongst speakers. In Figure 2 we show that the dental topic encompasses 28.66\% of slides whereas psychology encompasses a much smaller portion of 1.53\% of slides. In addition, most topics fall under science and math, leaving humanities unrepresented in this dataset. Similarly, quantitative figures may not be adequately represented as tables and equations represented only 8.2\% of the dataset. Further studies on using mouse traces as input signals must be done. Speakers may not consistently use mouse traces, leading to some slides with stronger alignment and some slides with weaker alignment. Finally, this dataset does not encompass other miscellaneous information a speaker might present during their lecture such as animations, speech tone, or extraneous information presented through the form of videos, websites, virtual whiteboards, or other redirected sites.

\vspace{-1mm}
\textbf{Broader impacts}: There may be downstream effects in training models exclusively on this dataset, since content in humanities may not be equally represented. Social biases could also be encoded into the dataset based on the choice of images and content that speakers decide to include in their lectures, such as images with predominantly male representation or primarily English language.
We believe that \datasetshorts is a first step towards tackling multimodality and alignment in educational slides, and we aim to further expand it with diversity in speakers, languages, subjects, and lecture styles.

\vspace{-3mm}
\section{Conclusion}
\vspace{-2mm}

In conclusion, we present the \datasets as benchmark for developing AI technologies that can communicate multimodal knowledge in educational content. Our diversely sourced and richly annotated dataset contributes two challenging research tasks as a step towards educationally relevant goals: (1) automatic retrieval of spoken explanations given figures and (2) automatic retrieval of illustrative figures given spoken explanations.
Through benchmarking existing and newly proposed models, we outline future research directions in tackling weak crossmodal alignment, novel visual mediums, technical language, and long-range sequences to bring us closer towards intelligent and accessible tutoring aids.

\newpage

\bibliographystyle{plain}
\bibliography{sample}

\newpage 

\appendix

\section{Descriptions of Previous Lecture Datasets}
\label{asec:prev_lec}

\textbf{LectureBank Dataset} \cite{li2019should} is a manually-collected dataset of lecture slides, consisting of 1352 online lecture pdf files from 60 courses in Computer Science in 5 sub-domains: Machine Learning, NLP, DL, and IR. The dataset is annotated for each lecture's topic and  prerequisite relation topics based on taxonomy from \cite{fabbri2018tutorialbank}. This data, does not contain aligned transcripts and was used to predict prerequisite relations for a given lecture slide.

\textbf{ALV} \cite{galanopoulos2019temporal} is a lecture video dataset of artificially-generated lectures, where transcripts from lectures are randomly split in fragments then assembled by combining (stitching) exactly 20 randomly selected fragments from various videos. The resulting dataset only consists of transcripts. This work was developed for the purpose of evaluating lecture video fragmentation techniques.

\textbf{VLEngagement} \cite{bulathwela2020vlengagement}, is dataset which was designed to study engagement in video lectures, where content-based (stop-word counts) and video-specific features (silence, video duration) are extracted from publicly available scientific video lectures

\textbf{LectureVideoDB} \cite{lectureVideoDB2018} is a dataset consisting of 5000 frames of lecture videos, with annotated text characters developed for the purposed of text detection and recognition in Lecture Videos.

\textbf{GoogleI/O} \cite{chen2014multi} is a dataset consisting of 209 presentation videos from the Google I/O conferences in the years 2010-2012. In this dataset, the authors offer only textual information from the speech and the slides. The retrieval task is done at the video level, where entire transcripts are matched with all the text in a presentation.

\textbf{LaRochelle} \cite{van2014multi} 47 French lecture recordings from author's lab, Similar to \cite{chen2014multi}, the authors study video-level retrieval. In addition, the authors experiment with cross-modal retrieval where a bag of words approach is used for the text and visual tokens. 

\section{MTurk: Annotators}
\label{asec:annotator_misc}

For each task, we approximate the time each takes with internal annotators to ensure a minimum payment of \$8 per hour. For the task of slide segmentation, as annotators are simply required to scroll through the video to find transition points, we pay 50 cents for a 15 minute long video (i.e \$2 for an hour long video). We pay annotators a total amount of \$856.95 for this task. For the task of figure annotations, we pay the annotators 5 cents per slide, where annotators are expected to spend around 20 seconds per slide. As a result, we spent \$451.55 for a total of 9031 slides.


\newpage
\section{MTurk: Slide Segmentation}
\label{asec:mturk_slideseg}
\begin{figure*}[!htb]
    \begin{center}
    \vspace{-4mm}
    \makebox[\textwidth][c]{\includegraphics[width=1\textwidth]{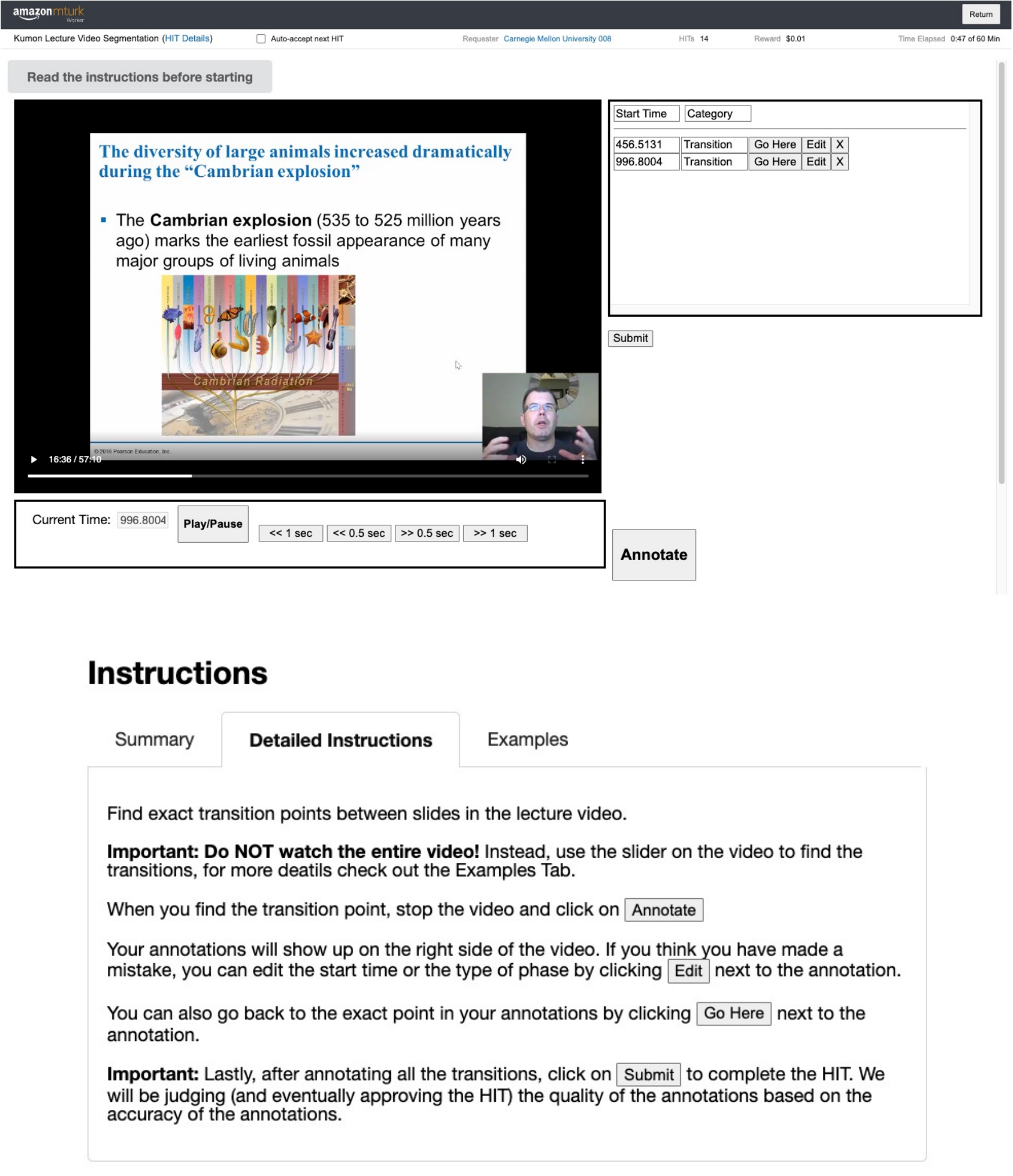}}
    \end{center}
    \caption{MTurk Screenshot and Instructions for Slide Segmentation}
\end{figure*}

\newpage
\section{MTurk: Figure Annotation}
\label{asec:mturk_figanot}
\begin{figure*}[!htb]
    \begin{center}
    \vspace{-4mm}
    \makebox[\textwidth][c]{\includegraphics[width=0.9\textwidth]{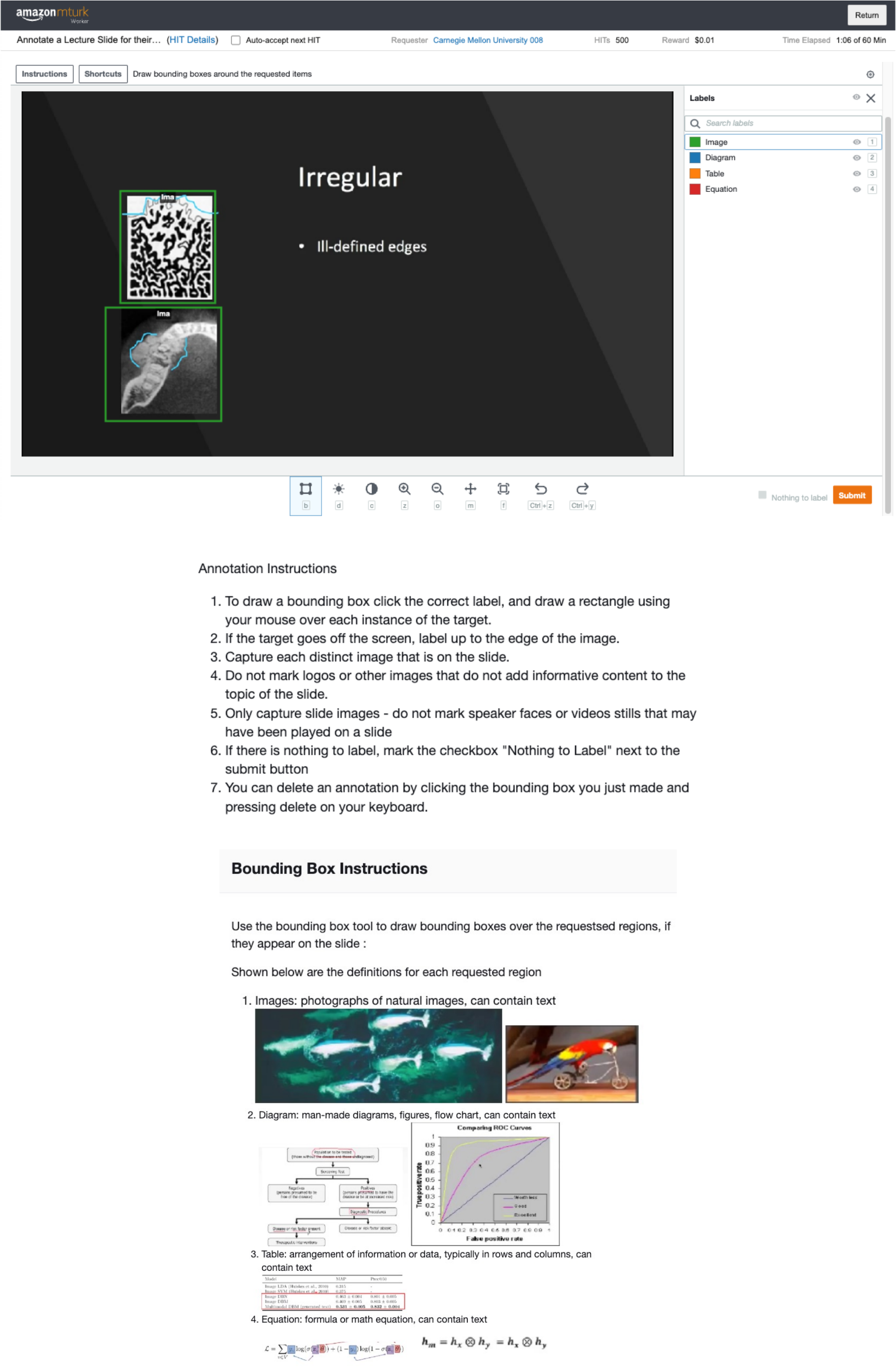}}
    \end{center}
    \caption{MTurk Screenshot and Instructions for Figure Annotations}
\end{figure*}

\newpage
\section{Training Details}
\label{asec:training_deets}
We use PyTorch as the auto-differentiation library to train all our models. For each speaker, with split the data such that a random 80\% is used as training data and the remaining 20\% is used for test (the data is split according to each random seed). In our experiments, we use the following hyperparameters. We train for 100 epochs, and our batch size is 8. 

We also utilize the 3 losses (MIL with a margin parameter $\lambda_m$, Diversity  $\lambda_{div}$, Domain Discrepancy $\lambda_{dom}$) as motivated in \cite{song2019polysemous}, we refer the audience to the original paper for the formulation of these losses. We use the default parameters, $\lambda_m = 0.1$, $\lambda_{div} = 0.01$, $\lambda_{dom}= 0.01$. For the number of locally guided features $K$ as shown in Figure \ref{fig:model}, we use $K=5$. Further finetuning on these hyperparameters is a future direction of study to boost performance. 

As mentioned in Section \ref{sec:polyvilt}, we use a pre-trained backbone ViLT encoder from HuggingFace, by the original authors, which has been trained on masked language modelling and image-text matching ('ViLT-b32-mlm-itm') \cite{wolf2019huggingface, kim2021ViLT}. We will release the full code base with our default hyperparameters.

The models were trained with CMU Multicomp Lab's internal cluster. The average model train runtime was around 8 hours on Titan X 1080 GPUs.

\newpage
\section{Comprehensive Results for Each Speaker}
\begin{table}[H]
\begin{tabular}{@{}llllllll@{}}
\toprule
\multicolumn{1}{c}{\multirow{2}{*}{\textbf{anat-1}}} & \multicolumn{3}{c}{Figure-to-Text}                                    & \textbf{} & \multicolumn{3}{c}{Text-to-Figure}                                    \\ \cmidrule(lr){2-4} \cmidrule(l){6-8} 
\multicolumn{1}{c}{}                                 & Recall@1              & Recall@5              & Recall@10             & \textbf{} & Recall@1              & Recall@5              & Recall@10             \\ \midrule
Random                                               & 0.27 ± 0.38           & 3.18 ± 2.4            & 5.77 ± 1.73           &           & 0.26 ± 0.36           & 3.38 ± 1.42           & 8.07 ± 3.8            \\
CLIP                                                 & 0.53 ± 0.37           & 4.53 ± 1.21           & 9.11 ± 1.11           &           & 0.54 ± 0.38           & 3.19 ± 1.63           & 7.17 ± 3.15           \\
PVSE                                                 & 2.1 ± 0.4             & 7.33 ± 1.14           & 12.76 ± 2.07          &           & 1.83 ± 0.36           & 8.09 ± 0.92           & 11.73 ± 0.88          \\
PVSE (BERT)                                          & 2.62 ± 0.43           & 5.49 ± 0.12           & 10.48 ± 1.71          &           & 1.04 ± 0.36           & 7.01 ± 2.21           & 10.68 ± 1.82          \\
PCME                                                 & 1.3 ± 0.71            & 4.18 ± 0.32           & 7.83 ± 0.16           &           & 1.3 ± 0.71            & 4.18 ± 0.32           & 7.83 ± 0.16           \\
PCME (BERT)                                          & 1.3 ± 0.71            & 3.92 ± 0.63           & 8.09 ± 0.92           &           & 1.3 ± 0.71            & 3.92 ± 0.63           & 8.09 ± 0.92           \\ \midrule
Ours                                                 & \textbf{11.23 ± 0.91} & 30.82 ± 6.1           & 42.31 ± 4.82          &           & \textbf{13.79 ± 2.34} & 34.34 ± 8.91          & 44.9 ± 6.53           \\
Ours w/ Trace                                        & 9.64 ± 3.08           & \textbf{31.05 ± 6.71} & \textbf{46.49 ± 2.67} &           & 10.71 ± 0.54          & \textbf{36.86 ± 2.33} & \textbf{49.85 ± 1.14} \\ \bottomrule
\end{tabular}
\end{table}

\begin{table}[H]
\begin{tabular}{@{}llllllll@{}}
\toprule
\multicolumn{1}{c}{\multirow{2}{*}{\textbf{anat-2}}} & \multicolumn{3}{c}{Figure-to-Text}                                   & \textbf{} & \multicolumn{3}{c}{Text-to-Figure}                                   \\ \cmidrule(lr){2-4} \cmidrule(l){6-8} 
\multicolumn{1}{c}{}                                 & Recall@1             & Recall@5              & Recall@10             & \textbf{} & Recall@1             & Recall@5              & Recall@10             \\ \midrule
Random                                               & 1.75 ± 2.48          & 21.05 ± 8.59          & 50.87 ± 9.92          &           & 8.77 ± 2.48          & 31.58 ± 8.6           & 56.14 ± 8.94          \\
CLIP                                                 & 7.12 ± 2.42          & 23.2 ± 6.48           & 57.21 ± 7.01          &           & 3.51 ± 4.96          & 16.08 ± 8.61          & 37.43 ± 3.61          \\
PVSE                                                 & 7.02 ± 2.48          & 38.6 ± 2.48           & 66.67 ± 2.48          &           & 7.02 ± 2.48          & 38.6 ± 6.56           & 68.42 ± 0.0           \\
PVSE (BERT)                                          & 5.26 ± 4.3           & 33.33 ± 4.96          & 61.4 ± 8.94           &           & 5.26 ± 0.0           & 33.34 ± 6.56          & 52.63 ± 8.59          \\
PCME                                                 & 5.26 ± 0.0           & 28.07 ± 4.96          & 56.14 ± 2.48          &           & 5.26 ± 0.0           & 28.07 ± 4.96          & 56.14 ± 2.48          \\
PCME (BERT)                                          & 5.26 ± 0.0           & 28.07 ± 2.48          & 52.63 ± 0.0           &           & 5.26 ± 0.0           & 28.07 ± 2.48          & 52.63 ± 0.0           \\ \midrule
Ours                                                 & \textbf{7.02 ± 2.48} & \textbf{54.39 ± 6.56} & \textbf{78.95 ± 4.3}  &           & \textbf{8.77 ± 2.48} & \textbf{57.89 ± 4.3}  & 75.44 ± 6.56          \\
Ours w/ Trace                                        & \textbf{8.77 ± 4.96} & \textbf{49.12 ± 6.56} & \textbf{73.68 ± 7.44} &           & 7.02 ± 6.56          & \textbf{49.12 ± 8.94} & \textbf{77.19 ± 6.56} \\ \bottomrule
\end{tabular}
\end{table}
\begin{table}[H]
\begin{tabular}{@{}llllllll@{}}
\toprule
\multicolumn{1}{c}{\multirow{2}{*}{\textbf{bio-1}}} & \multicolumn{3}{c}{Figure-to-Text}                                  & \textbf{} & \multicolumn{3}{c}{Text-to-Figure}                                   \\ \cmidrule(lr){2-4} \cmidrule(l){6-8} 
\multicolumn{1}{c}{}                                & Recall@1             & Recall@5             & Recall@10             & \textbf{} & Recall@1             & Recall@5              & Recall@10             \\ \midrule
Random                                              & 0.79 ± 0.43          & 3.13 ± 1.19          & 4.7 ± 0.89            &           & 8.77 ± 2.48          & 31.58 ± 8.6           & 56.14 ± 8.94          \\
CLIP                                                & 0.51 ± 0.03          & 3.41 ± 1.53          & 5.7 ± 2.4             &           & 3.51 ± 4.96          & 16.08 ± 8.61          & 37.43 ± 3.61          \\
PVSE                                                & 0.97 ± 0.07          & 4.05 ± 0.54          & 6.0 ± 1.06            &           & 7.02 ± 2.48          & 38.6 ± 6.56           & 68.42 ± 0.0           \\
PVSE (BERT)                                         & 0.79 ± 0.43          & 5.07 ± 1.15          & 8.55 ± 1.52           &           & 5.26 ± 0.0           & 33.34 ± 6.56          & 52.63 ± 8.59          \\
PCME                                                & 0.66 ± 0.29          & 2.11 ± 0.4           & 4.68 ± 0.49           &           & 5.26 ± 0.0           & 28.07 ± 4.96          & 56.14 ± 2.48          \\
PCME (BERT)                                         & 0.48 ± 0.03          & 2.39 ± 0.23          & 4.68 ± 0.49           &           & 5.26 ± 0.0           & 28.07 ± 2.48          & 52.63 ± 0.0           \\ \midrule
Ours                                                & \textbf{4.23 ± 0.9}  & \textbf{12.53 ± 2.6} & \textbf{19.15 ± 1.29} &           & \textbf{8.77 ± 2.48} & \textbf{57.89 ± 4.3}  & 75.44 ± 6.56          \\
Ours w/ Trace                                       & \textbf{2.91 ± 0.82} & \textbf{7.14 ± 1.08} & \textbf{12.65 ± 2.73} &           & 7.02 ± 6.56          & \textbf{49.12 ± 8.94} & \textbf{77.19 ± 6.56} \\ \bottomrule
\end{tabular}
\end{table}
\begin{table}[H]
\begin{tabular}{@{}llllllll@{}}
\toprule
\multicolumn{1}{c}{\multirow{2}{*}{\textbf{bio-3}}} & \multicolumn{3}{c}{Figure-to-Text}                                   & \textbf{} & \multicolumn{3}{c}{Text-to-Figure}                                   \\ \cmidrule(lr){2-4} \cmidrule(l){6-8} 
\multicolumn{1}{c}{}                                & Recall@1             & Recall@5              & Recall@10             & \textbf{} & Recall@1             & Recall@5              & Recall@10             \\ \midrule
Random                                              & 0.57 ± 0.4           & 3.68 ± 1.63           & 6.58 ± 1.56           &           & 1.16 ± 1.11          & 4.07 ± 1.72           & 8.35 ± 1.33           \\
CLIP                                                & 0.0 ± 0.0            & 5.4 ± 1.52            & 11.35 ± 2.29          &           & 0.85 ± 0.02          & 3.66 ± 1.04           & 7.09 ± 1.22           \\
PVSE                                                & 1.14 ± 0.81          & 6.61 ± 0.93           & 15.8 ± 1.33           &           & 1.16 ± 0.43          & 6.34 ± 1.25           & 12.35 ± 1.28          \\
PVSE (BERT)                                         & 2.87 ± 1.11          & 7.47 ± 0.94           & 12.93 ± 1.45          &           & 1.43 ± 0.39          & 5.12 ± 1.02           & 9.19 ± 0.7            \\
PCME                                                & 1.7 ± 0.65           & 5.15 ± 0.56           & 8.28 ± 2.13           &           & 1.7 ± 0.65           & 5.15 ± 0.56           & 8.28 ± 2.13           \\
PCME (BERT)                                         & 1.7 ± 0.65           & 5.44 ± 0.76           & 9.16 ± 1.55           &           & 1.7 ± 0.65           & 5.44 ± 0.76           & 9.16 ± 1.55           \\ \midrule
Ours                                                & \textbf{1.74 ± 0.75} & \textbf{12.03 ± 0.37} & \textbf{19.14 ± 1.56} &           & \textbf{4.57 ± 1.45} & \textbf{13.11 ± 3.02} & \textbf{20.04 ± 2.47} \\
Ours w/ Trace                                       & \textbf{1.17 ± 0.83} & \textbf{8.85 ± 1.9}   & \textbf{14.85 ± 3.03} &           & 3.42 ± 0.6           & \textbf{10.03 ± 0.35} & \textbf{16.36 ± 2.08} \\ \bottomrule
\end{tabular}
\end{table}
\begin{table}[H]
\begin{tabular}{@{}llllllll@{}}
\toprule
\multicolumn{1}{c}{\multirow{2}{*}{\textbf{bio-4}}} & \multicolumn{3}{c}{Figure-to-Text}                                  & \textbf{} & \multicolumn{3}{c}{Text-to-Figure}                                  \\ \cmidrule(lr){2-4} \cmidrule(l){6-8} 
\multicolumn{1}{c}{}                                & Recall@1             & Recall@5              & Recall@10            & \textbf{} & Recall@1            & Recall@5              & Recall@10             \\ \midrule
Random                                              & 0.77 ± 0.44          & 2.15 ± 0.45           & 4.62 ± 1.04          &           & 0.77 ± 0.21         & 2.93 ± 0.98           & 5.84 ± 1.13           \\
CLIP                                                & 0.32 ± 0.46          & 2.48 ± 0.16           & 4.95 ± 0.83          &           & 0.16 ± 0.23         & 1.88 ± 0.81           & 5.16 ± 1.5            \\
PVSE                                                & 1.7 ± 1.44           & 4.75 ± 1.41           & 7.82 ± 1.96          &           & 2.17 ± 1.18         & 4.31 ± 1.25           & 6.45 ± 1.62           \\
PVSE (BERT)                                         & 1.08 ± 0.58          & 2.61 ± 0.57           & 5.07 ± 0.09          &           & 1.22 ± 0.75         & 3.52 ± 0.72           & 5.66 ± 1.61           \\
PCME                                                & 1.86 ± 1.98          & 3.39 ± 1.55           & 4.92 ± 1.54          &           & 1.86 ± 1.98         & 3.39 ± 1.55           & 4.92 ± 1.54           \\
PCME (BERT)                                         & 0.62 ± 0.22          & 1.23 ± 0.57           & 3.99 ± 1.17          &           & 0.62 ± 0.22         & 1.23 ± 0.57           & 3.99 ± 1.17           \\ \midrule
Ours                                                & \textbf{4.28 ± 2.04} & \textbf{12.22 ± 6.66} & \textbf{18.5 ± 9.12} &           & \textbf{2.9 ± 1.72} & \textbf{11.79 ± 5.52} & \textbf{20.1 ± 6.1}   \\
Ours w/ Trace                                       & \textbf{3.67 ± 1.82} & \textbf{12.07 ± 5.46} & \textbf{19.9 ± 6.82} &           & 2.29 ± 0.97         & \textbf{12.68 ± 5.64} & \textbf{21.88 ± 7.02} \\ \bottomrule
\end{tabular}
\end{table}
\begin{table}[H]
\begin{tabular}{@{}llllllll@{}}
\toprule
\multicolumn{1}{c}{\multirow{2}{*}{\textbf{dental}}} & \multicolumn{3}{c}{Figure-to-Text}                                 & \textbf{} & \multicolumn{3}{c}{Text-to-Figure}                                 \\ \cmidrule(lr){2-4} \cmidrule(l){6-8} 
\multicolumn{1}{c}{}                                 & Recall@1             & Recall@5             & Recall@10            & \textbf{} & Recall@1             & Recall@5             & Recall@10            \\ \midrule
Random                                               & 0.17 ± 0.14          & 0.87 ± 0.25          & 1.91 ± 0.28          &           & 0.29 ± 0.21          & 0.92 ± 0.29          & 1.67 ± 0.28          \\
CLIP                                                 & 0.06 ± 0.08          & 1.09 ± 0.35          & 2.14 ± 0.25          &           & 0.23 ± 0.08          & 0.98 ± 0.23          & 1.85 ± 0.32          \\
PVSE                                                 & 0.4 ± 0.21           & 1.73 ± 0.13          & 2.48 ± 0.42          &           & 0.29 ± 0.08          & 1.44 ± 0.19          & 2.65 ± 0.26          \\
PVSE (BERT)                                          & 0.34 ± 0.0           & 1.84 ± 0.57          & 2.65 ± 0.33          &           & 0.64 ± 0.31          & 1.57 ± 0.65          & 2.6 ± 0.32           \\
PCME                                                 & 0.23 ± 0.08          & 0.86 ± 0.13          & 1.73 ± 0.41          &           & 0.23 ± 0.08          & 0.86 ± 0.13          & 1.73 ± 0.41          \\
PCME (BERT)                                          & 0.23 ± 0.08          & 0.86 ± 0.23          & 1.67 ± 0.34          &           & 0.23 ± 0.08          & 0.86 ± 0.23          & 1.67 ± 0.34          \\ \midrule
Ours                                                 & \textbf{0.63 ± 0.29} & \textbf{2.72 ± 0.23} & \textbf{6.18 ± 0.53} &           & \textbf{1.15 ± 0.16} & \textbf{5.31 ± 0.25} & \textbf{8.36 ± 1.45} \\
Ours w/ Trace                                        & \textbf{0.69 ± 0.23} & \textbf{3.28 ± 1.04} & \textbf{6.16 ± 1.1}  &           & 0.8 ± 0.39           & \textbf{3.28 ± 0.69} & \textbf{5.88 ± 0.58} \\ \bottomrule
\end{tabular}
\end{table}
\begin{table}[H]
\begin{tabular}{@{}llllllll@{}}
\toprule
\multicolumn{1}{c}{\multirow{2}{*}{\textbf{ml-1}}} & \multicolumn{3}{c}{Figure-to-Text}                                 & \textbf{} & \multicolumn{3}{c}{Text-to-Figure}                                 \\ \cmidrule(lr){2-4} \cmidrule(l){6-8} 
\multicolumn{1}{c}{}                               & Recall@1             & Recall@5             & Recall@10            & \textbf{} & Recall@1             & Recall@5             & Recall@10            \\ \midrule
Random                                             & 0.28 ± 0.2           & 1.88 ± 0.28          & 3.5 ± 0.48           &           & 0.29 ± 0.21          & 0.92 ± 0.29          & 1.67 ± 0.28          \\
CLIP                                               & 0.43 ± 0.34          & 1.69 ± 0.36          & 4.83 ± 1.94          &           & 0.23 ± 0.08          & 0.98 ± 0.23          & 1.85 ± 0.32          \\
PVSE                                               & 1.48 ± 0.47          & 5.65 ± 0.66          & 7.51 ± 0.96          &           & 0.29 ± 0.08          & 1.44 ± 0.19          & 2.65 ± 0.26          \\
PVSE (BERT)                                        & 0.54 ± 0.16          & 3.78 ± 1.36          & 6.44 ± 1.18          &           & 0.64 ± 0.31          & 1.57 ± 0.65          & 2.6 ± 0.32           \\
PCME                                               & 0.66 ± 0.34          & 2.43 ± 0.19          & 4.49 ± 0.61          &           & 0.23 ± 0.08          & 0.86 ± 0.13          & 1.73 ± 0.41          \\
PCME (BERT)                                        & 0.54 ± 0.16          & 2.68 ± 0.53          & 4.61 ± 0.48          &           & 0.23 ± 0.08          & 0.86 ± 0.23          & 1.67 ± 0.34          \\ \midrule
Ours                                               & \textbf{0.82 ± 0.05} & \textbf{4.76 ± 1.93} & \textbf{7.89 ± 1.87} &           & \textbf{1.15 ± 0.16} & \textbf{5.31 ± 0.25} & \textbf{8.36 ± 1.45} \\
Ours w/ Trace                                      & \textbf{1.22 ± 0.3}  & \textbf{3.71 ± 0.89} & \textbf{6.2 ± 1.84}  &           & 0.8 ± 0.39           & \textbf{3.28 ± 0.69} & \textbf{5.88 ± 0.58} \\ \bottomrule
\end{tabular}
\end{table}
\begin{table}[H]
\begin{tabular}{@{}llllllll@{}}
\toprule
\multicolumn{1}{c}{\multirow{2}{*}{\textbf{psy-1}}} & \multicolumn{3}{c}{Figure-to-Text}                                   & \textbf{} & \multicolumn{3}{c}{Text-to-Figure}                                   \\ \cmidrule(lr){2-4} \cmidrule(l){6-8} 
\multicolumn{1}{c}{}                                & Recall@1             & Recall@5              & Recall@10             & \textbf{} & Recall@1             & Recall@5              & Recall@10             \\ \midrule
Random                                              & 4.48 ± 4.78          & 11.1 ± 6.21           & 21.22 ± 2.19          &           & 4.15 ± 1.35          & 15.36 ± 1.05          & 26.73 ± 1.25          \\
CLIP                                                & 4.05 ± 3.89          & 13.22 ± 3.03          & 30.03 ± 5.88          &           & 2.68 ± 2.34          & 13.38 ± 2.66          & 22.35 ± 2.92          \\
PVSE                                                & 3.99 ± 0.86          & 16.71 ± 4.75          & 29.65 ± 4.62          &           & 5.07 ± 2.48          & 17.98 ± 1.51          & 35.8 ± 1.29           \\
PVSE (BERT)                                         & \textbf{5.71 ± 2.87} & 18.87 ± 3.56          & 27.79 ± 3.23          &           & 4.16 ± 1.39          & 20.46 ± 3.25          & 34.84 ± 2.7           \\
PCME                                                & 5.08 ± 2.49          & 14.75 ± 1.36          & 26.29 ± 3.24          &           & 4.16 ± 1.39          & 15.68 ± 2.66          & 30.92 ± 9.63          \\
PCME (BERT)                                         & 4.16 ± 1.39          & 13.54 ± 6.14          & 26.93 ± 10.49         &           & 4.16 ± 1.39          & 14.46 ± 7.45          & 26.93 ± 10.49         \\ \midrule
Ours                                                & \textbf{2.27 ± 3.21} & \textbf{19.5 ± 0.76}  & \textbf{38.23 ± 2.51} &           & \textbf{9.38 ± 5.52} & \textbf{32.4 ± 1.45}  & \textbf{43.52 ± 5.59} \\
Ours w/ Trace                                       & \textbf{3.39 ± 1.54} & \textbf{19.18 ± 3.81} & \textbf{33.78 ± 4.26} &           & 7.51 ± 3.75          & \textbf{20.83 ± 6.17} & \textbf{37.8 ± 5.6}   \\ \bottomrule
\end{tabular}
\end{table}
\begin{table}[H]
\begin{tabular}{@{}llllllll@{}}
\toprule
\multicolumn{1}{c}{\multirow{2}{*}{\textbf{psy-2}}} & \multicolumn{3}{c}{Figure-to-Text}                                   & \textbf{} & \multicolumn{3}{c}{Text-to-Figure}                                   \\ \cmidrule(lr){2-4} \cmidrule(l){6-8} 
\multicolumn{1}{c}{}                                & Recall@1             & Recall@5              & Recall@10             & \textbf{} & Recall@1             & Recall@5              & Recall@10             \\ \midrule
Random                                              & 4.48 ± 4.78          & 11.1 ± 6.21           & 21.22 ± 2.19          &           & 0.44 ± 0.62          & 5.96 ± 2.95           & 12.74 ± 4.36          \\
CLIP                                                & 4.05 ± 3.89          & 13.22 ± 3.03          & 30.03 ± 5.88          &           & 1.62 ± 1.46          & 5.77 ± 1.85           & 14.12 ± 0.88          \\
PVSE                                                & 3.99 ± 0.86          & 16.71 ± 4.75          & 29.65 ± 4.62          &           & 3.47 ± 2.08          & 11.2 ± 2.34           & 19.22 ± 1.49          \\
PVSE (BERT)                                         & \textbf{5.71 ± 2.87} & 18.87 ± 3.56          & 27.79 ± 3.23          &           & \textbf{4.47 ± 0.6}  & 11.56 ± 1.24          & 18.26 ± 2.92          \\
PCME                                                & 5.08 ± 2.49          & 14.75 ± 1.36          & 26.29 ± 3.24          &           & 2.18 ± 2.24          & 8.91 ± 3.03           & 17.98 ± 2.88          \\
PCME (BERT)                                         & 4.16 ± 1.39          & 13.54 ± 6.14          & 26.93 ± 10.49         &           & 1.83 ± 0.76          & 8.63 ± 3.26           & 15.94 ± 6.21          \\ \midrule
Ours                                                & \textbf{2.27 ± 3.21} & \textbf{19.5 ± 0.76}  & \textbf{38.23 ± 2.51} &           & \textbf{1.36 ± 1.08} & \textbf{14.89 ± 7.3}  & \textbf{26.92 ± 5.74} \\
Ours w/ Trace                                       & \textbf{3.39 ± 1.54} & \textbf{19.18 ± 3.81} & \textbf{33.78 ± 4.26} &           & 2.72 ± 2.15          & \textbf{16.06 ± 0.29} & \textbf{27.66 ± 2.79} \\ \bottomrule
\end{tabular}
\end{table}
\begin{table}[H]
\begin{tabular}{@{}llllllll@{}}
\toprule
\multicolumn{1}{c}{\multirow{2}{*}{\textbf{speaking}}} & \multicolumn{3}{c}{Figure-to-Text}                                    & \textbf{} & \multicolumn{3}{c}{Text-to-Figure}                                   \\ \cmidrule(lr){2-4} \cmidrule(l){6-8} 
\multicolumn{1}{c}{}                                   & Recall@1              & Recall@5              & Recall@10             & \textbf{} & Recall@1             & Recall@5              & Recall@10             \\ \midrule
Random                                                 & 3.26 ± 2.72           & 21.46 ± 6.17          & 44.79 ± 7.36          &           & 0.44 ± 0.62          & 5.96 ± 2.95           & 12.74 ± 4.36          \\
CLIP                                                   & 4.34 ± 1.65           & 14.16 ± 7.03          & 38.77 ± 3.21          &           & 1.62 ± 1.46          & 5.77 ± 1.85           & 14.12 ± 0.88          \\
PVSE                                                   & 8.54 ± 1.64           & 27.64 ± 5.15          & 51.18 ± 5.45          &           & 3.47 ± 2.08          & 11.2 ± 2.34           & 19.22 ± 1.49          \\
PVSE (BERT)                                            & \textbf{7.43 ± 1.39}  & 24.44 ± 2.67          & 38.4 ± 3.71           &           & \textbf{4.47 ± 0.6}  & 11.56 ± 1.24          & 18.26 ± 2.92          \\
PCME                                                   & 3.19 ± 0.1            & 16.04 ± 3.08          & 32.01 ± 3.53          &           & 2.18 ± 2.24          & 8.91 ± 3.03           & 17.98 ± 2.88          \\
PCME (BERT)                                            & 3.19 ± 0.1            & 15.97 ± 0.49          & 30.83 ± 0.59          &           & 1.83 ± 0.76          & 8.63 ± 3.26           & 15.94 ± 6.21          \\ \midrule
Ours                                                   & \textbf{13.75 ± 3.68} & \textbf{29.58 ± 7.24} & \textbf{53.06 ± 4.52} &           & \textbf{1.36 ± 1.08} & \textbf{14.89 ± 7.3}  & \textbf{26.92 ± 5.74} \\
Ours w/ Trace                                          & \textbf{5.28 ± 2.9}   & \textbf{32.71 ± 9.07} & \textbf{52.92 ± 9.48} &           & \textbf{2.72 ± 2.15} & \textbf{16.06 ± 0.29} & \textbf{27.66 ± 2.79} \\ \bottomrule
\end{tabular}
\caption{Speaker-wise results for Figure-to-Text and Text-to-Figure retrieval. PolyViLT consistently outperforms previous baselines. }
\newpage
\end{table}

\label{asec:each_sp_results}

\newpage
\section{Keyword Identifiability}

\begin{table}[H]
\resizebox{0.5\textwidth}{!}{%
\begin{tabular}{@{}lrrrr@{}}
\toprule
\multirow{2}{*}{\textbf{\begin{tabular}[c]{@{}l@{}}PolyViLT\\ r@10\end{tabular}}} & \multicolumn{4}{c}{tfidf rank} \\ \cmidrule(l){2-5} 
 & \multicolumn{1}{l}{\textless 5} & \multicolumn{1}{l}{5 -10} & \multicolumn{1}{l}{10 - 30} & \multicolumn{1}{l}{30 - 50} \\ \midrule
Text-to-Figure & \textbf{0.236} & 0.2 & 0.122 & 0.132 \\
Figure-to-Text & \textbf{0.249} & 0.22 & 0.066 & 0.124 \\ \bottomrule
\end{tabular}%
}
\caption{Recall@10 scores for Keyword Identifiability measured by TF-IDF ranks}
\label{tab:tfidf}
\end{table}

Specifically for figures which contain text, which consists of 54.9\% of our dataset, there are many cases where the pairing between text and figures can be easily found by identifying the keyword and finding its existence in the figure or the spoken language. Naively finding the existence of identical words in two instances is trivial and could lead to incorrect retrievals. The core challenge lies in correctly identifying the keyword that defines the slide segment. 

In order to understand the importance of identifying the keyword and how our model performs for text-inclusive figures, we measure the term frequency–inverse document frequency (or tf-idf) of each word in the spoken language, except stopwords which are filtered out. The words are then ranked according to their tf-idf values. We iterate through each word, find the words that also exist the ocr output of the figure and extract the word with the lowest tf-idf rank. Under this condition, if the tf-idf rank for a word is 5, this can be intuitively seen as the fifth most important keyword that defines the slide. Simply stated, if the tf-idf rank of a word is low, the keyword can be easily detected in the slide and the spoken language. On the other hand, if the tf-idf rank of a word is high, this implies that the keyword is hard to detect. \footnote{Note that this method of retrieval is intractable with more number of words and documents }

In Table \ref{tab:tfidf}  We measure the recall@10 score conditioned on tf-idf ranks, which indicates how well PolyViLT does under varying levels of difficulty of identifying the keyword. PolyViLT's struggles for cases with easier keyword identifiability and suffers even more with harder cases. This calls for a need for PolyViLT to effectively address easier cases, via using tf-idf directly as a feature, and relying more on the vision when the keyword is not easily identifiable.

\newpage
\section{Long Range Sequence and OOV Tokens}
\label{asec:language_problems}
\begin{table}[H]
\begin{tabular}{@{}lllllrlllll@{}}
\toprule
\multicolumn{1}{c}{\textbf{CLIP}} & \multicolumn{5}{l}{\textbf{(a) Length of Spoken Language}} & \textbf{} & \multicolumn{4}{l}{\textbf{(b) Number of Subwords}} \\ \cmidrule(l){2-11} 
\multicolumn{1}{c}{\textbf{r@10}} & \textless{}100 & 100 - 200 & 200 - 400 & 400 - 600 & 600+  &           & \textless{}10    & 10 - 20   & 20-30    & 30 - 50   \\ \midrule
Figure-to-Text                    & 0.0447         & 0.0465    & 0.0567    & 0.0676    & 0.175 &           & 0.065            & 0.062     & 0.0543   & 0.0619    \\
Text-to-Figure                    & 0.0793         & 0.0662    & 0.0599    & 0.0571    & 0.14  &           & 0.0704           & 0.055     & 0.0498   & 0.0473    \\ \bottomrule
\end{tabular}
\end{table}
\begin{table}[H]
\begin{tabular}{@{}lllllllllll@{}}
\toprule
\multicolumn{1}{c}{\textbf{PVSE}} & \multicolumn{5}{l}{\textbf{(a) Length of Spoken Language}}                    & \textbf{} & \multicolumn{4}{l}{\textbf{(b) Number of Subwords}}                                                  \\ \cmidrule(l){2-11} 
\multicolumn{1}{c}{\textbf{r@10}} & \textless{}100 & 100 - 200 & 200 - 400 & 400 - 600 & \multicolumn{1}{r}{600+} &           & \textless{}10 & 10 - 20                    & 20-30                      & 30 - 50                    \\ \midrule
Figure-to-Text                    & 0.0779         & 0.0777    & 0.0644    & 0.0901    & 0.0928                   &           & 0.0973        & \multicolumn{1}{r}{0.0842} & \multicolumn{1}{r}{0.0656} & \multicolumn{1}{r}{0.0667} \\
Text-to-Figure                    & 0.0901         & 0.116     & 0.1063    & 0.1013    & 0.0814                   &           & 0.108         & 0.0839                     & 0.0602                     & 0.084                      \\ \bottomrule
\end{tabular}
\end{table}
\begin{table}[H]
\begin{tabular}{@{}lllllllllll@{}}
\toprule
\multicolumn{1}{c}{\textbf{PCME}} & \multicolumn{5}{l}{\textbf{(a) Length of Spoken Language}}                    & \textbf{} & \multicolumn{4}{l}{\textbf{(b) Number of Subwords}} \\ \cmidrule(l){2-11} 
\multicolumn{1}{c}{\textbf{r@10}} & \textless{}100 & 100 - 200 & 200 - 400 & 400 - 600 & \multicolumn{1}{r}{600+} &           & \textless{}10    & 10 - 20   & 20-30    & 30 - 50   \\ \midrule
Figure-to-Text                    & 0.0342         & 0.1301    & 0.082     & 0.0733    & 0.053                    &           & 0.0744           & 0.0556    & 0.0617   & 0.0271    \\
Text-to-Figure                    & 0.0342         & 0.1301    & 0.082     & 0.0752    & 0.0536                   &           & 0.076            & 0.0518    & 0.0603   & 0.0309    \\ \bottomrule
\end{tabular}

\caption{For all competitive baselines, performance (a) peaks at 100-200 words then drops with increasing length of spoken language and (b) drops with increasing number of subwords}
\end{table}

\newpage
\section{Qualitative Cases of Failure}
\label{asec:qualitative}
\begin{figure*}[!htb]
    \begin{center}
    \vspace{-4mm}
    \makebox[\textwidth][c]{\includegraphics[width=1\textwidth]{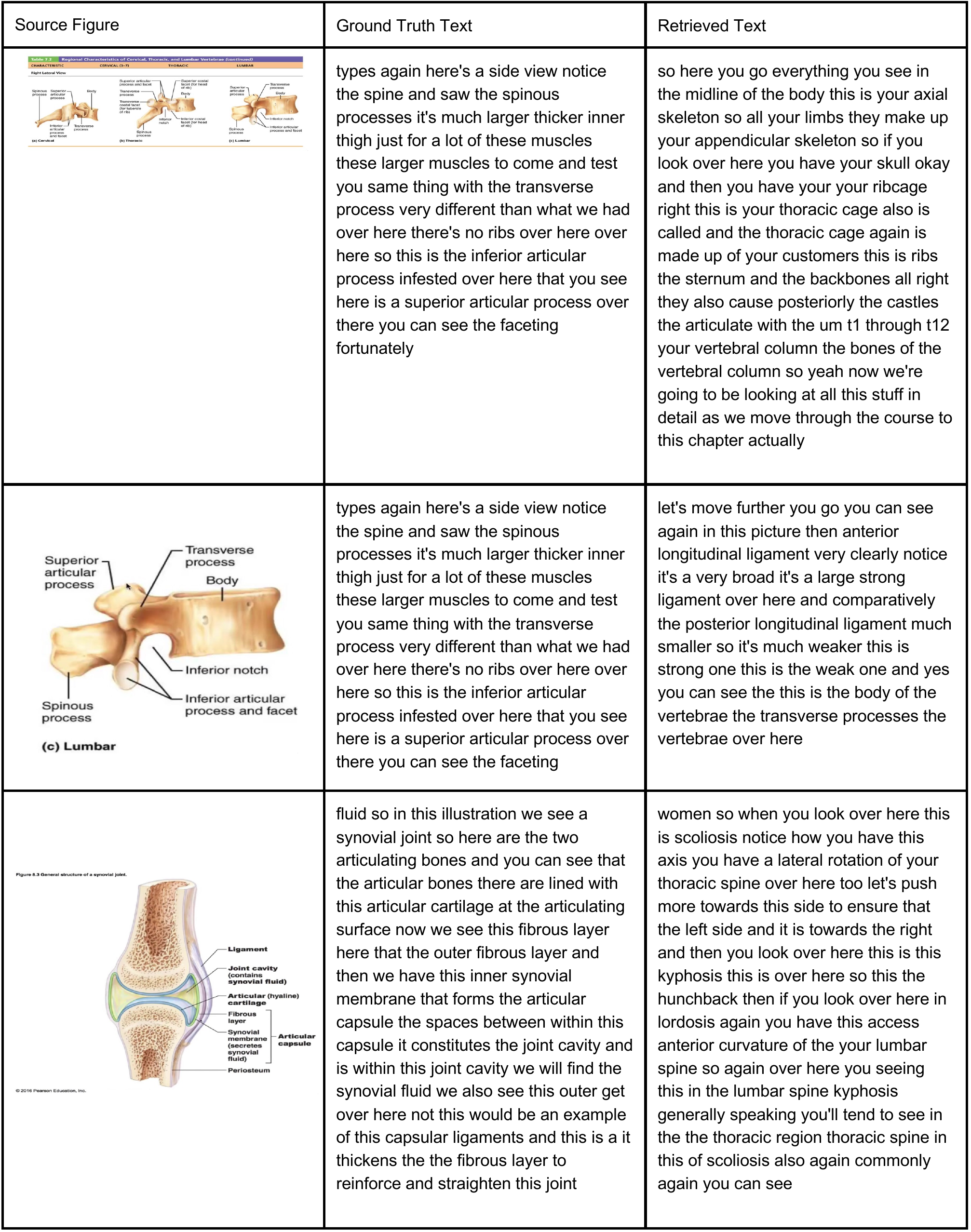}}
    \end{center}
    \caption{Figure-to-Text: Failure Case for Anat-1 (top-1 retrieval result shown on right)}
\end{figure*}

\begin{figure*}[!htb]
    \begin{center}
    \vspace{-4mm}
    \makebox[\textwidth][c]{\includegraphics[width=1\textwidth]{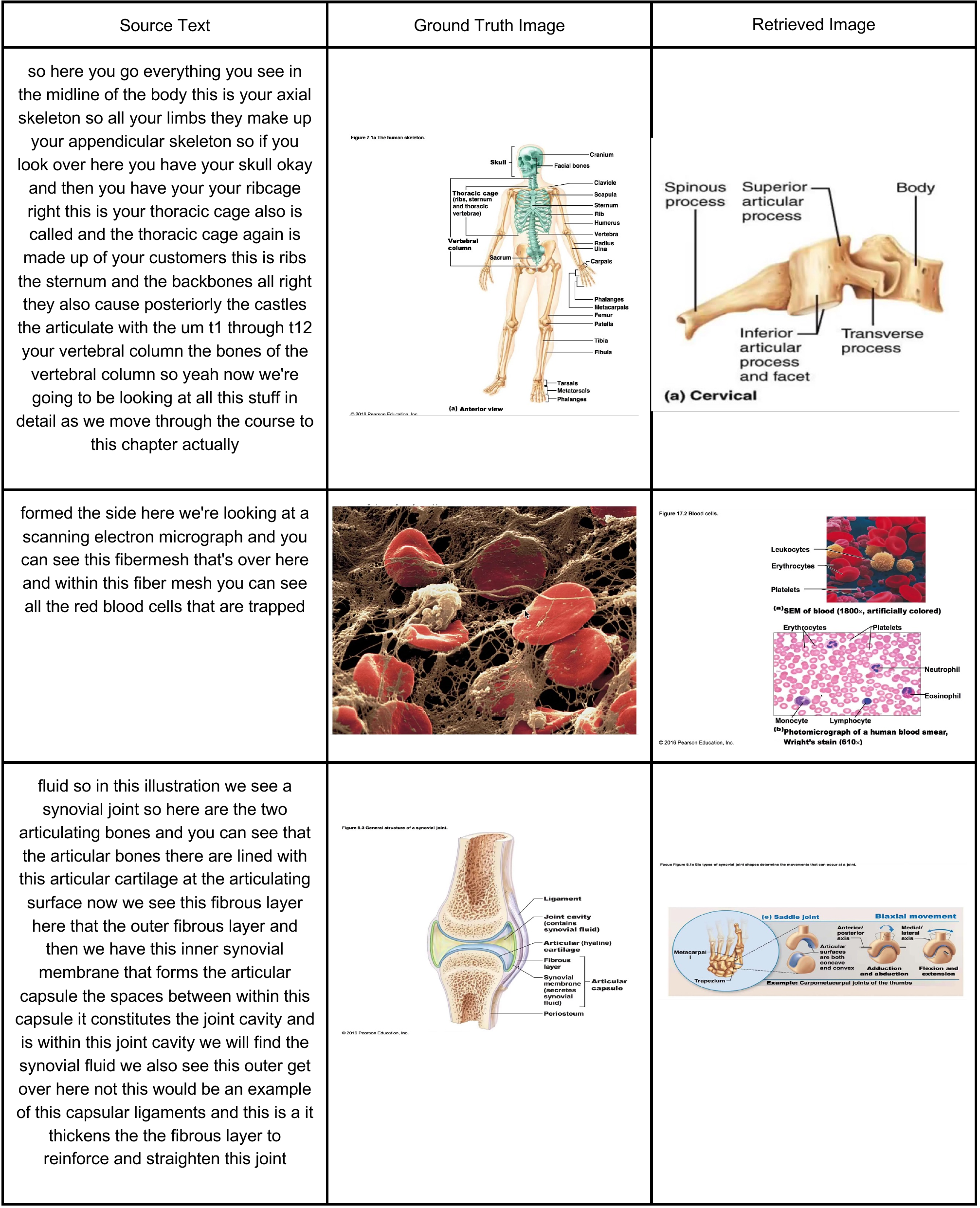}}
    \end{center}
    \caption{Text-to-Figure: Failure Cases for Anat-1 (top-1 retrieval result shown on right)}
\end{figure*}

\begin{figure*}[!htb]
    \begin{center}
    \vspace{-4mm}
    \makebox[\textwidth][c]{\includegraphics[width=1\textwidth]{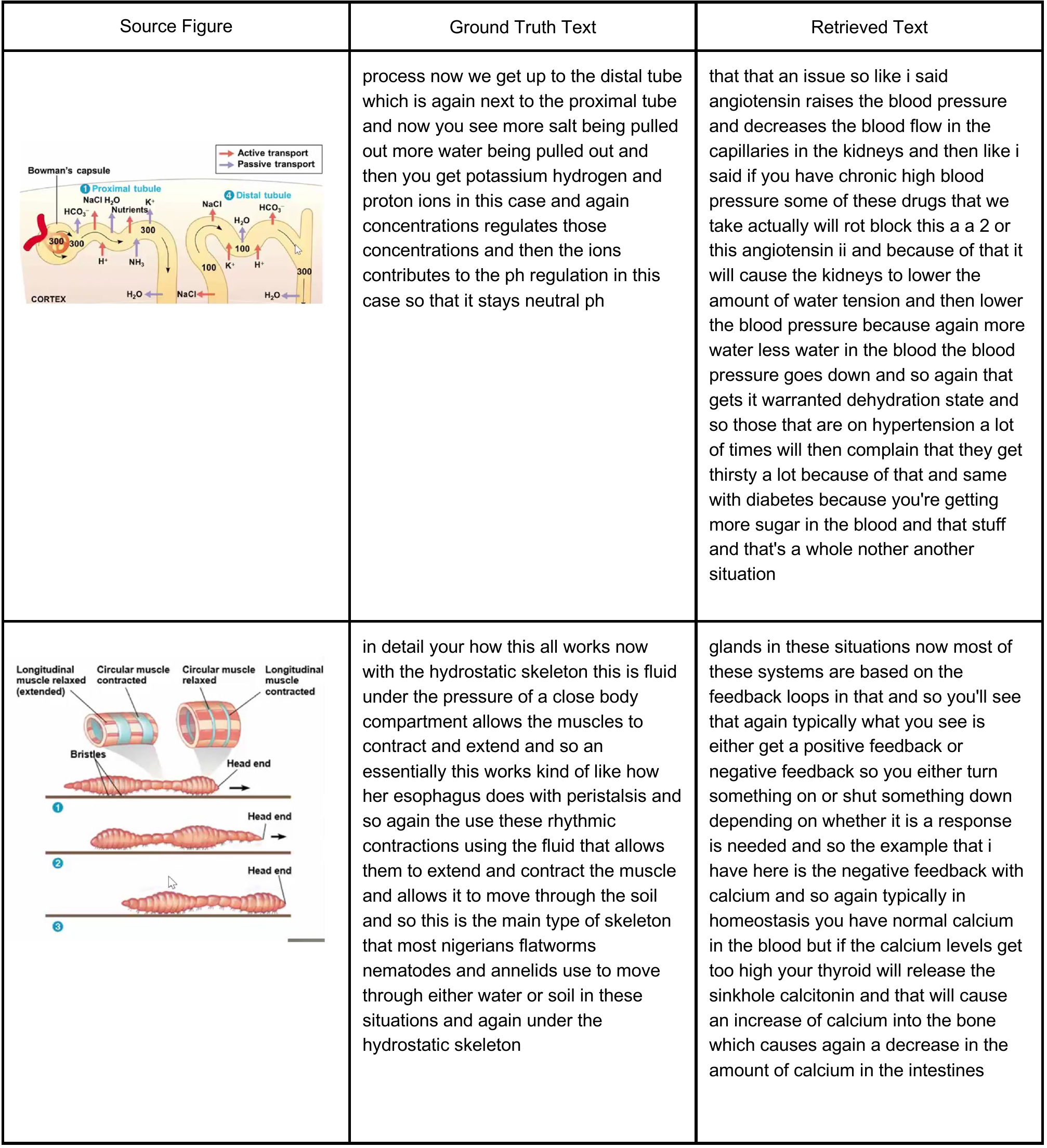}}
    \end{center}
    \caption{Figure-to-Text: Failure Case for Bio-1 (top-1 retrieval result shown on right)}
\end{figure*}

\begin{figure*}[!htb]
    \begin{center}
    \vspace{-4mm}
    \makebox[\textwidth][c]{\includegraphics[width=1\textwidth]{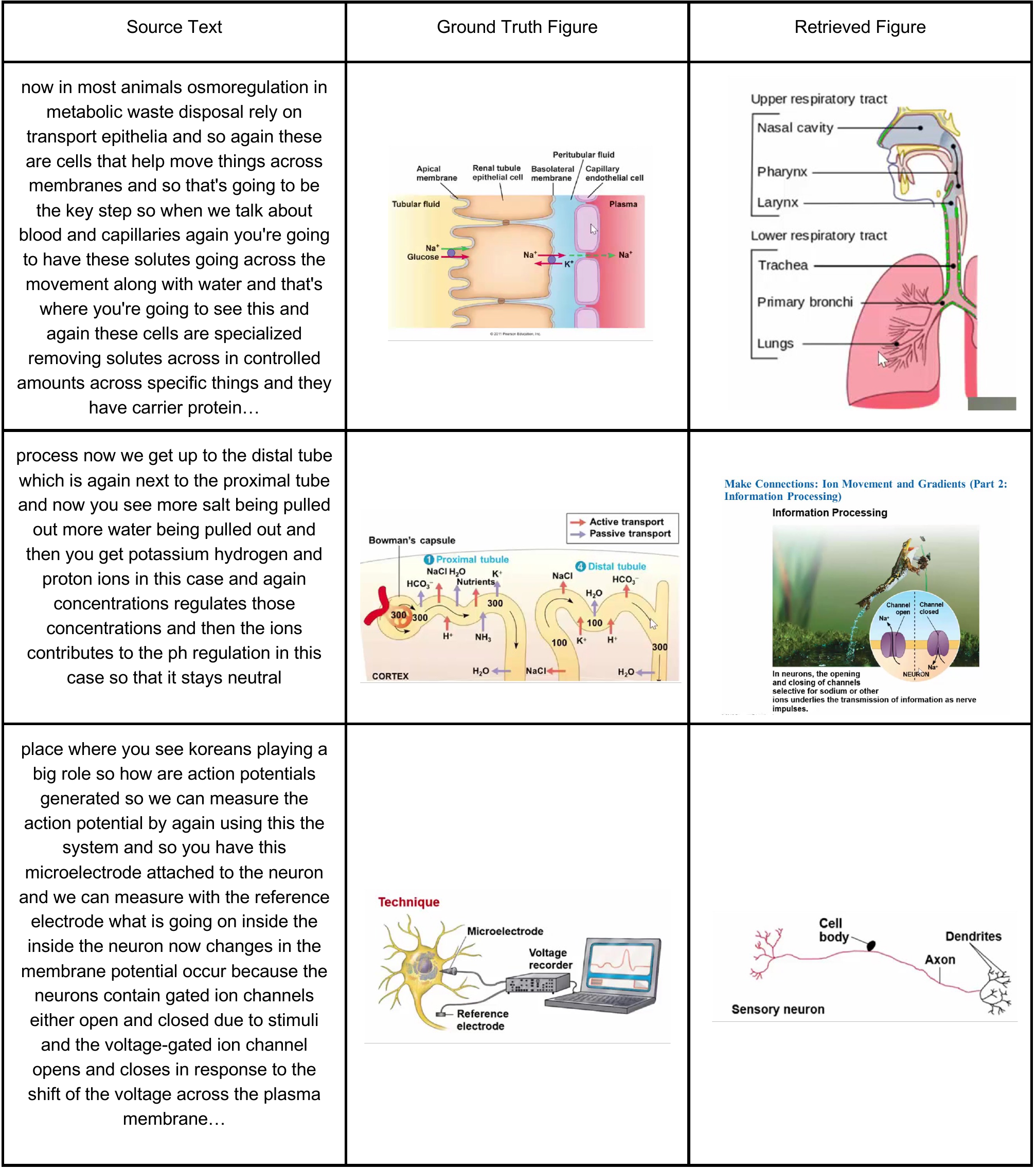}}
    \end{center}
    \caption{Text-to-Figure: Failure Cases for Bio-1 (top-1 retrieval result shown on right)}
\end{figure*}

\begin{figure*}[!htb]
    \begin{center}
    \vspace{-4mm}
    \makebox[\textwidth][c]{\includegraphics[width=1\textwidth]{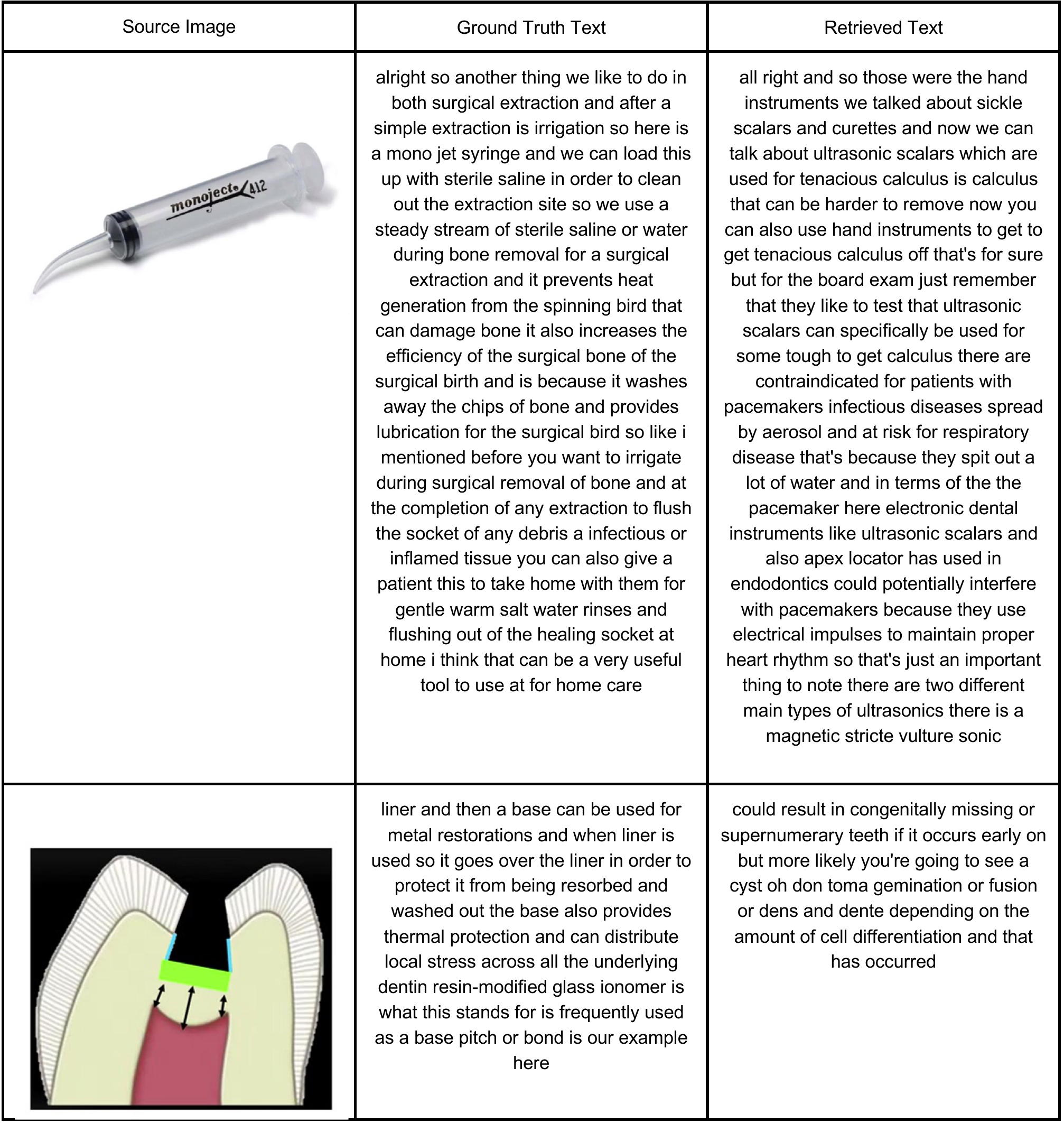}}
    \end{center}
    \caption{Figure-to-Text: Failure Case for dental (top-1 retrieval result shown on right)}
\end{figure*}

\begin{figure*}[!htb]
    \begin{center}
    \vspace{-4mm}
    \makebox[\textwidth][c]{\includegraphics[width=1\textwidth]{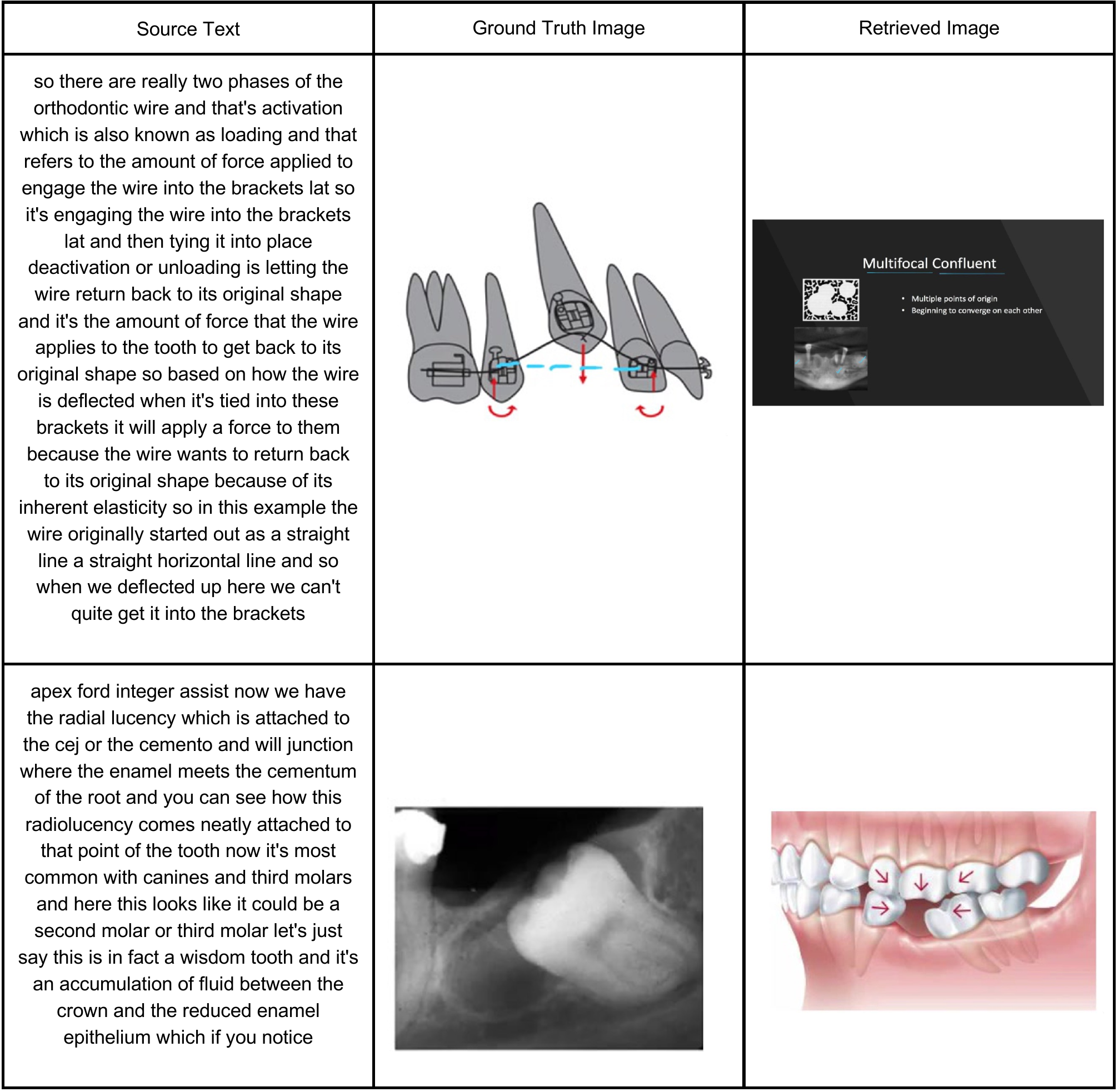}}
    \end{center}
    \caption{Text-to-Figure: Failure Cases for dental (top-1 retrieval result shown on right)}
\end{figure*}

\begin{figure*}[!htb]
    \begin{center}
    \vspace{-4mm}
    \makebox[\textwidth][c]{\includegraphics[width=1\textwidth]{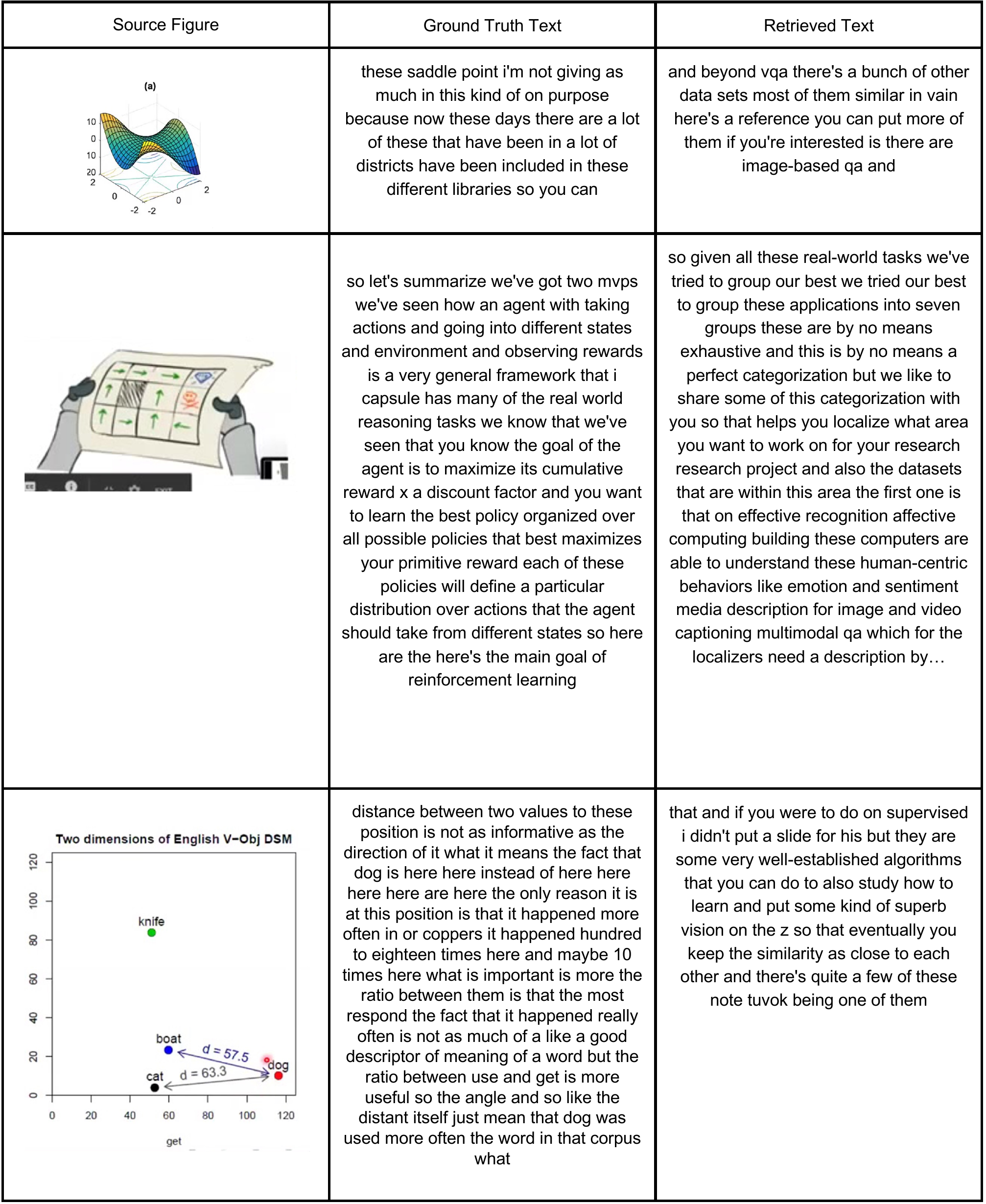}}
    \end{center}
    \caption{Figure-to-Text: Failure Case for ml-1 (top-1 retrieval result shown on right)}
\end{figure*}

\begin{figure*}[!htb]
    \begin{center}
    \vspace{-4mm}
    \makebox[\textwidth][c]{\includegraphics[width=1\textwidth]{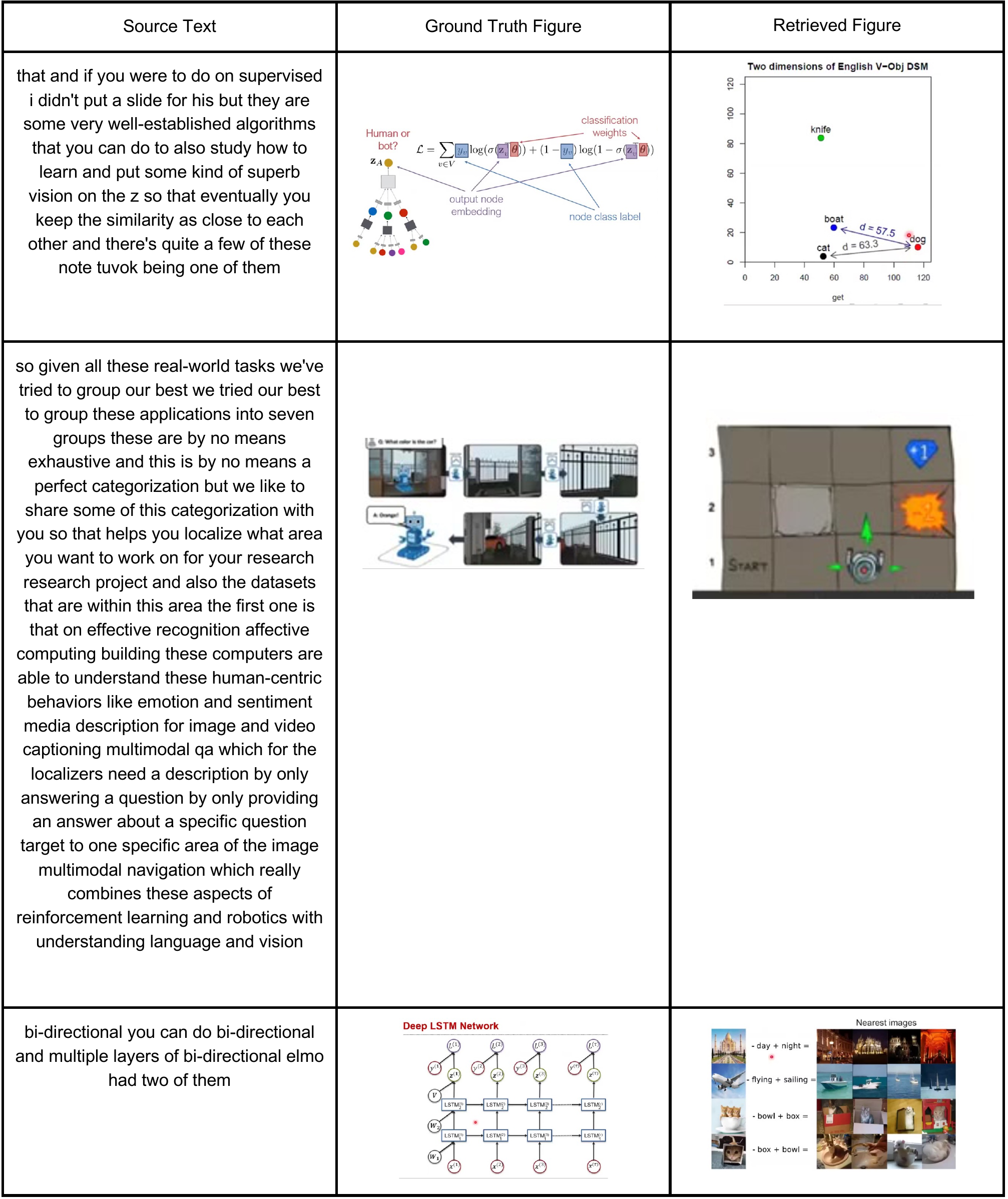}}
    \end{center}
    \caption{Text-to-Figure: Failure Cases for ml-1 (top-1 retrieval result shown on right)}
\end{figure*}

\begin{figure*}[!htb]
    \begin{center}
    \vspace{-4mm}
    \makebox[\textwidth][c]{\includegraphics[width=1\textwidth]{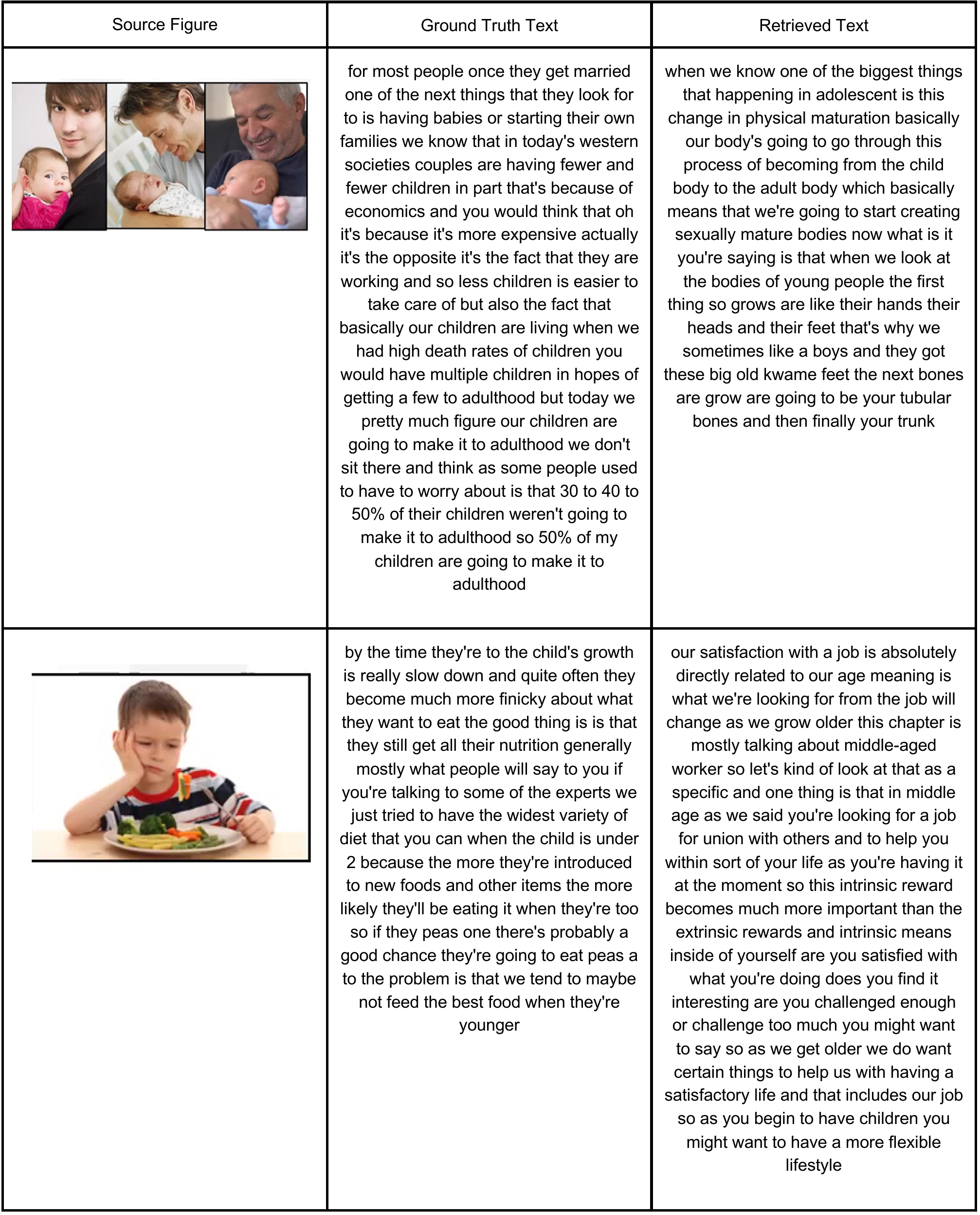}}
    \end{center}
    \caption{Figure-to-Text: Failure Case for Psy-2 (top-1 retrieval result shown on right)}
\end{figure*}

\begin{figure*}[!htb]
    \begin{center}
    \vspace{-4mm}
    \makebox[\textwidth][c]{\includegraphics[width=1\textwidth]{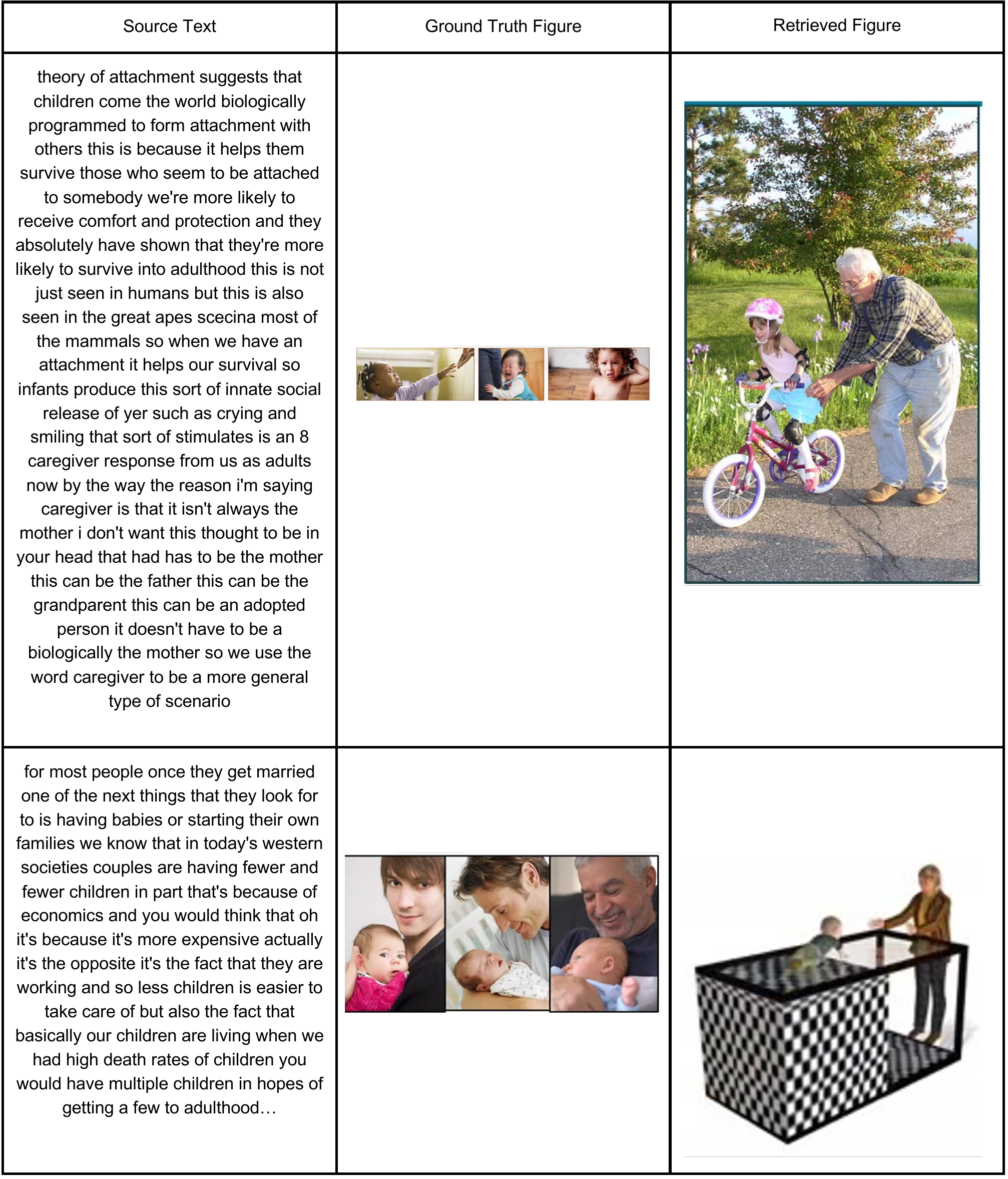}}
    \end{center}
    \caption{Text-to-Figure: Failure Cases for Psy-2 (top-1 retrieval result shown on right)}
\end{figure*}

\end{document}